\documentclass[letterpaper]{article} 
\usepackage{aaai2026}  
\usepackage{times}  
\usepackage{helvet}  
\usepackage{courier}  
\usepackage[hyphens]{url}  
\usepackage{graphicx} 
\usepackage{natbib}  
\usepackage{caption} 
\usepackage{algorithm}
\usepackage{algorithmic}

\usepackage{booktabs}       
\usepackage{amsmath,amssymb}
\usepackage{multirow}
\usepackage{subcaption}
\usepackage{xcolor}

\usepackage{newfloat}
\usepackage{listings}

\newtheorem{theorem}{Theorem}

\newtheorem{definition}[theorem]{Definition}

\usepackage{pifont}
\usepackage{amsmath,amssymb}

\usepackage{color}

\usepackage[table,xcdraw]{xcolor}
\definecolor{mygray}{gray}{0.88}

\def\1{\bm{1}}

\usepackage{fontawesome5}  
\usepackage{xcolor}

\usepackage{booktabs}
\usepackage{multirow}

\usepackage{mathrsfs}
\usepackage{makecell}

\def\rvx{{\mathbf{x}}}

\DeclareCaptionStyle{ruled}{labelfont=normalfont,labelsep=colon,strut=off} 
\floatstyle{ruled}
\newfloat{listing}{tb}{lst}{}
\floatname{listing}{Listing}
\title{Finding Time Series Anomalies using Granular-ball Vector Data Description}
\author {
    Lifeng Shen\textsuperscript{\rm 1},
    Liang Peng\textsuperscript{\rm 1},
    Ruiwen Liu\textsuperscript{\rm 1},
    Shuyin Xia\textsuperscript{\rm 1}\thanks{Corresponding Author.},
    Yi Liu\textsuperscript{\rm 2}
}
\affiliations {
    \textsuperscript{\rm 1}Chongqing Key Laboratory of Computational Intelligence, Key Laboratory of Cyberspace Big Data Intelligent Security, Ministry of Education, Sichuan-Chongqing Co-construction Key Laboratory of Digital Economy Intelligence and Key Laboratory of Big Data Intelligent Computing, Chongqing University of Posts and Telecommunications\\
    \textsuperscript{\rm 2}Chongqing Ant Consumer Finance Co,. Ltd , Ant Group.\\
    shenlf@cqupt.edu.cn, l1angpeng@foxmail.com, liuruiwen622@foxmail.com, xiasy@cqupt.edu.cn, larry.liuy@myxiaojin.cn
}

\begin{document}

\maketitle

\begin{abstract}
Modeling normal behavior in dynamic, nonlinear time series data is challenging for effective anomaly detection. Traditional methods, such as nearest neighbor and clustering approaches, often depend on rigid assumptions, such as a predefined number of reliable neighbors or clusters, which frequently break down in complex temporal  scenarios. 
To address these limitations, we introduce the Granular-ball One-Class Network (GBOC), a novel approach based on a data-adaptive representation called Granular-ball Vector Data Description (GVDD). 
GVDD partitions the latent space into compact, high-density regions represented by granular-balls, which are generated through a density-guided hierarchical splitting process and refined by removing noisy structures. Each granular-ball serves as a prototype for local normal behavior, naturally positioning itself between individual instances and clusters while preserving the local topological structure of the sample set.  
During training, GBOC improves the compactness of representations by aligning samples with their nearest granular-ball centers. During inference, anomaly scores are computed based on the distance to the nearest granular-ball. 
By focusing on dense, high-quality regions and significantly reducing the number of prototypes, GBOC delivers both robustness and efficiency in anomaly detection. Extensive experiments validate the effectiveness and superiority of the proposed method, highlighting its ability to handle the challenges of time series anomaly detection. The source code of the paper is available at https://github.com/notshine/GBOC.
\end{abstract}


\section{Introduction}
Complex cyber-physical systems, such as power plants, data centers, and smart factories, rely on a multitude of sensors operating concurrently to continuously generate high-volume multivariate time series data. 
Timely and accurate detection of anomalies within such data streams is critical, as it enables effective monitoring of system health, prevention of potential failures, and reduction of operational risks. 
Time series anomaly detection, involves identifying deviations from expected behavior at the level of individual time steps, with diverse real-world manifestations \cite{Schmidl2022Anomaly}. Examples of such anomalous events include physical attacks on industrial systems \cite{SWAT}, unpredictable robot behavior \cite{Robot}, faulty sensors from wide-sensor networks \cite{faulty1, faulty2}, cybersecurity attacks on server monitoring systems \cite{OmniAnomaly, cybersecurity2}, and spacecraft malfunctions observed via telemetry sensor data \cite{spacecraft3}.

Describing normal data in time series is crucial for effective anomaly detection. Commonly used approaches, such as nearest neighbor and clustering methods, aim to characterize normal patterns. Nearest neighbor approaches, like KNN \cite{SubKNN}, detect anomalies based on proximity. However, they struggle with the dynamic and complex nature of time series data. For instance, in group anomalies, a set of abnormal points may still have enough ``neighbors" within the group, making them appear normal. This issue arises because these methods rely on local density or proximity, failing to capture the global or temporal structure of the data.

Clustering-based methods, such as KShapeAD \cite{KshapeAD1, KshapeAD2, SAND}, 
group data points into clusters representing normal behavior. These approaches generally assume that normal data forms dense, well-defined clusters, while anomalies are located far from clusters or in sparse, low-density regions. One-class classification methods, such as OCSVM \cite{OCSVM} and SVDD \cite{SVDD}, further simplify this concept by modeling normal data as a single cluster, often enclosed within a hypersphere. 
To enhance the flexibility of normality modeling, DeepSVDD \cite{deepsvdd} leverages deep learning to learn a more adaptive hypersphere, while THOC \cite{THOC} introduces a multiscale vector data description through a differentiable hierarchical clustering process. Similarly, memory-based approaches, such as MEMTO \cite{MEMTO}, extend clustering by incorporating a memory module to store representative prototypes (or centroids) of normal clusters.

\begin{figure}[t!]
    \centering  \includegraphics[width=0.8\linewidth]{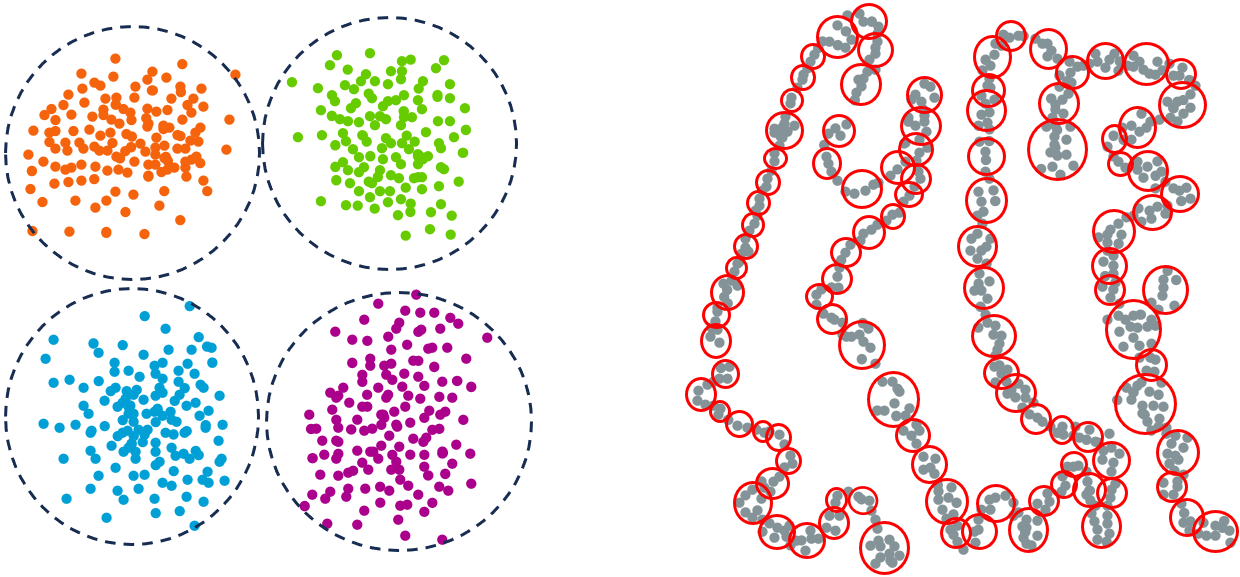}
    \caption{Latent space comparison between clustering-based modeling (left) and the proposed method preserving structural continuity (right). Clustering-based methods assume well-defined boundaries, which are unsuitable for the continuous and boundary-free distributions typical of time series data, requiring a more adaptive descriptive approach.}
    \label{fig:example_cluster}
\end{figure}

While clustering approaches are conceptually intuitive, they face notable limitations. Most clustering methods require the number of clusters to be predefined, which is particularly problematic when the true data structure is unknown or involves complex, multimodal distributions. One-class classification methods address this by representing normal behavior with a single hypersphere, but this oversimplified assumption struggles to capture the diversity and complexity of multimodal patterns. Similarly, memory-based methods depend heavily on the quality and representativeness of stored prototypes. When these prototypes fail to encompass the full variability of normal behavior, the methods become biased and significantly less effective.

These limitations are especially pronounced in dynamic and nonlinear time series data. As illustrated in Figure \ref{fig:example_cluster} (right), time series sliding windows' representations often exhibit structural continuity, characterized by smooth transitions between patterns. In contrast, the rigid cluster assignments depicted in Figure \ref{fig:example_cluster} (left) rely on discrete boundaries and separability, making them ill-suited to handle the continuous and dynamic nature of such representations.

To address the challenges of modeling normal behavior in complex and dynamic time series data, we propose the Granular-ball One-Class Network (GBOC). Unlike traditional methods, which rely on rigid assumptions such as predefined clusters or reliable neighbors, GBOC introduces the Granular-ball Vector Data Description (GVDD), a flexible representation that adaptively partitions the latent space into compact, high-density regions called granular-balls, illustrated in Figure \ref{fig:example_cluster} (right). These regions are constructed through a density-guided hierarchical splitting process and refined by pruning noisy or diffuse areas, ensuring robust modeling of normal behavior. During training, GBOC aligns samples with their nearest granular-ball centers to enhance representation compactness. At inference, anomaly scores are efficiently computed based on the distance to the nearest granular-ball center, leveraging the dense, high-quality regions. By reducing the number of prototypes compared to training samples, GBOC achieves computational efficiency and robustness. Experiments on diverse time series benchmarks demonstrate that GBOC outperforms traditional and deep learning-based methods in accuracy, scalability, and efficiency, offering a practical solution for real-world anomaly detection tasks.

\noindent \textbf{Contribution.} We propose the Granular-ball One-Class Network (GBOC), a novel time series anomaly detection method that introduces Granular-ball Vector Data Description (GVDD), the first extension of one-class methods by incorporating granular-ball computing. This design enables a time series adaptive vector data description, effectively alleviating the limitations of nearest-neighbor and clustering-based approaches. Extensive experiments demonstrate the superior performance of GBOC against strong baselines.

\begin{figure*}[t!]
    \centering  \includegraphics[width=0.86\linewidth,height=3.0in]{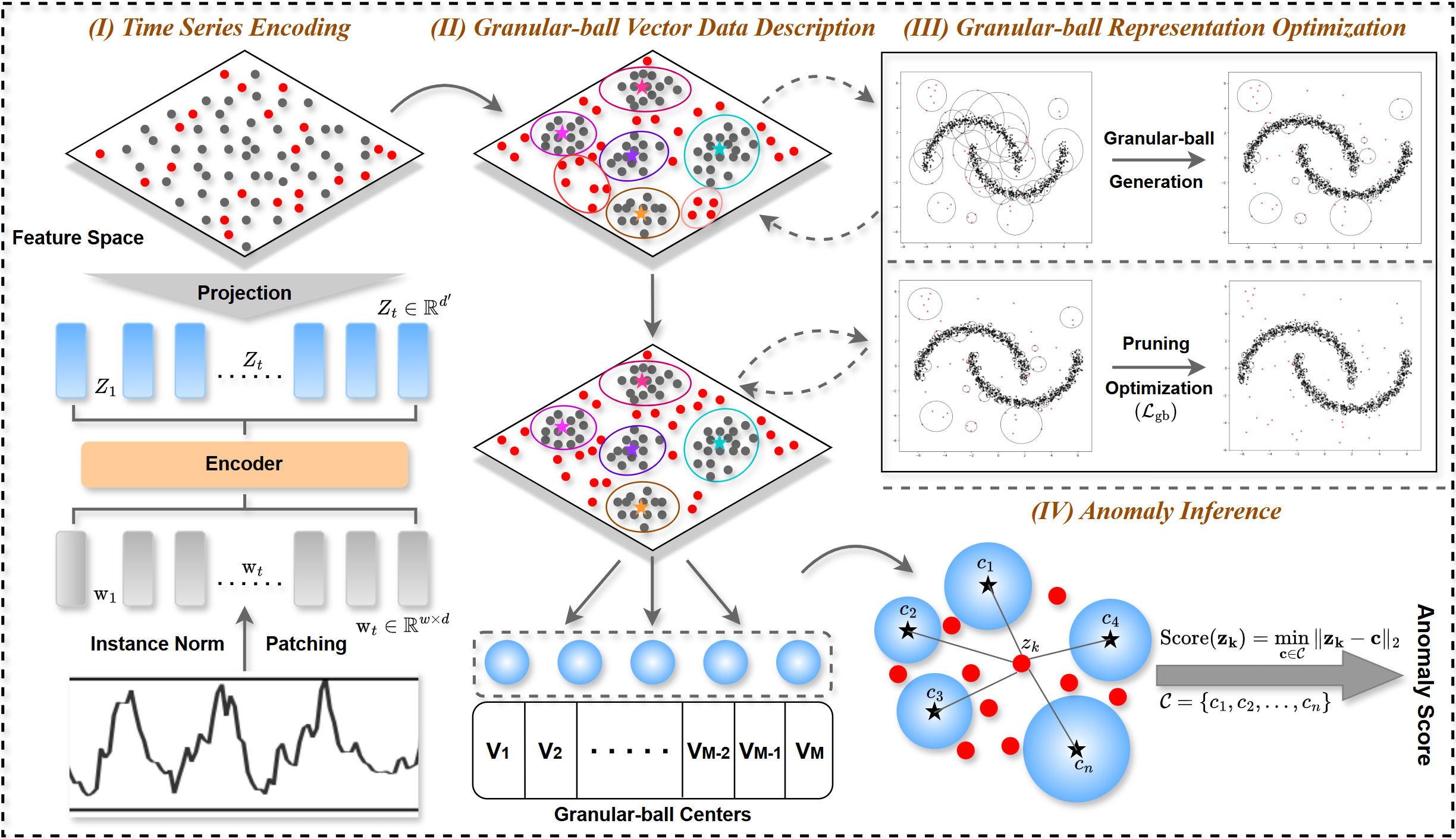}
    \caption{Illustration of the proposed Granular-ball One-Class Network (GBOC).}
    \label{fig:process}
\end{figure*}

\section{Preliminaries}\label{sec:pre}
\textbf{Problem Formulation:} 
A time series is a sequence of data points recorded at successive time intervals. It can be represented as:  
$\mathbf{X} = \begin{bmatrix} \mathbf{x}_1, \mathbf{x}_2, \ldots, \mathbf{x}_T \end{bmatrix}^\top \in \mathbb{R}^{T \times d}$,  
where $T$ is the number of time steps and $d$ is the number of channels. 
A time series anomaly is a data point $\mathbf{x}_t$ or subsequence $\mathbf{X}_{t_1:t_2} = [\mathbf{x}_{t_1}, \ldots, \mathbf{x}_{t_2}]^\top$ (for $1 \leq t_1 \leq t_2 \leq T$) that deviates significantly from expected patterns. Anomalies are rare and misaligned with the statistical properties of normal data.   
Time series anomaly detection aims to identify time series anomaly points that deviate significantly from the expected value $\hat{x}_t$ predicted by a detection model $\mathcal{M}$. For a predefined threshold $\delta_t > 0$, the detection criterion is:  
\(
\mathcal{M}(\rvx_t) > \delta_t
\), 
where $\mathcal{M}$ is trained to learn normal patterns from historical data and assigns an anomaly score to each observation. 
More related works on time series anomaly detection are provided in Appendix \ref{sec:related}.

\noindent \textbf{Granular-ball Computing (GBC)} \cite{GBC} provides a granular-ball computing (GBC) approach for modeling complex data distributions by abstracting them into adaptive local regions, called \emph{granular-balls}. This concept is inspired by the “global precedence” principle in cognitive science \cite{chen1982topological}, which states that human perception tends to prioritize global structural features over local details. Following this principle, GBC enables multi-granularity cognitive modeling \cite{wang2017dgcc}. Unlike traditional point-based models, GBC leverages region-based representations, offering compact, robust, and efficient data summarization. 
Formally, GBC represents the dataset $\{\rvx_t\}_{t=1}^n$ using a set of granular-balls, denoted by $GB = \{ GB_j \}_{j=1}^m$. The formal definition of a granular-ball is provided in Definition~\ref{defn:GB}.

\begin{definition}[Granular-ball (GB)]\label{defn:GB}
A \emph{granular-ball} $GB$ is a region in $\mathbb{R}^d$ characterized by its center $\mathbf{c} \in \mathbb{R}^d$ and radius $\mathbf{r} \in \mathbb{R}_{\geq 0}$, derived from a finite set of points $\{P_i\}_{i=1}^{|GB|} \subset \mathbb{R}^d$ contained within it.  
The center $\mathbf{c}$ is defined as the mean of all points in $GB$:  
$\mathbf{c} = \frac{1}{|GB|} \sum_{i=1}^{|GB|} \rvx_i$ 
and the radius $\mathbf{r}$ is defined as the maximum Euclidean distance from any point $\rvx_i \in GB$ to the center $\mathbf{c}$:  
\begin{equation}
    \mathbf{r} = \max_{\rvx_i \in GB} \| \rvx_i - \mathbf{c} \|.
\end{equation} 
\end{definition}

Based on the definition of a granular-ball, the construction of GBC follows a two-stage procedure. In the first stage, the dataset $\{\rvx_t\}_{t=1}^n$ is coarsely partitioned using K-Means into approximately $\sqrt{n}$ clusters, each forming an initial granular-ball. In the second stage, each granular-ball is recursively evaluated and refined into two child balls according to a density criterion: the \emph{granular-ball distribution measure (DM)}~\cite{xie2023efficient}, defined in Definition~\ref{def:GB_density}.
\begin{definition}[Granular-ball Distribution Measure] \label{def:GB_density}
Given a granular-ball $GB$ with center $\mathbf{c}$ and radius $\mathbf{r}$, the DM is defined by computing the ratio of the number $|GB|$ of data points and the sum radius $s$ in $GB$, as follows:  
\begin{equation}
\label{eq:DM}
DM(GB) = \frac{s}{|GB|},
\end{equation}  
where $s = \sum_{\mathbf{x} \in GB } \left \| \mathbf{x} - \mathbf{c} \right \|$. A smaller $DM$ value indicates better ball quality.  
\end{definition}

The granular-ball generation procedure is detailed in Algorithm \ref{alg:granular-ball}, provided in Appendix \ref{sec:GBC}. As observed, any granular-ball \( GB \) containing at least \( s_{\min} \) points is partitioned into two sub-balls by applying the 2-Means clustering algorithm to the points within \( GB \). Here, \( s_{\min} \) represents the minimum support requirement for a granular-ball to qualify for splitting, and it is empirically set to \( s_{\min} = 8 \). \begin{definition}[Refinement Criterion]\label{defn:Refinement} A granular-ball $GB$ is eligible for splitting if the weighted value of the decision metric (denoted as $DM$) after the split is strictly less than that of the original ball, i.e., $DM_w < DM(GB)$, and both resulting sub-balls satisfy the minimum support requirement. 
\end{definition}
The weighted decision metric after splitting is calculated as:  
\begin{equation}
\label{eq:DMW}
DM_w = \frac{|GB^{(1)}|}{|GB|} \, DM(GB^{(1)}) + \frac{|GB^{(2)}|}{|GB|} \, DM(GB^{(2)}),
\end{equation}
where $GB^{(1)}$ and $GB^{(2)}$ are the two sub-balls obtained by applying 2-Means clustering to the data points within $GB$. 
This recursive refinement process continues until no further quality-improving splits are possible \cite{W-GBC}. %
Upon termination, a set of granular-balls $\{ GB_j \}_{j=1}^m$ is produced to cover the dataset.

\section{Methodology}
In this section, we formally elaborate on the granular-ball one-class network (GBOC) for time series anomaly detection. As shown in Figure~\ref{fig:process}, GBOC operates in four steps: (i) encoding time series into a latent space, (ii) grouping normal patterns into high-density granular-balls, (iii) refining these patterns with pruning and alignment, and (iv) detecting anomalies by measuring the distance to the closest granular-ball center. Next, we will explain each part step by step.

\subsection{Time Series Encoding}
As a common practice, the input time series $\mathbf{X}$ is first segmented into overlapping windows $\{\mathbf{w}_1, \mathbf{w}_2, \dots, \mathbf{w}_{N_\text{w}}\}$ using a sliding window mechanism, where each window has length $w$. Each window $\mathbf{w}_i \in \mathbb{R}^{w \times d}$ is encoded into a $d'$-dimensional latent representation  
\begin{align}
   \mathbf{z}_i = f_\theta(\mathbf{w}_i),  
\end{align}  
producing the latent feature set \( Z = \{\mathbf{z}_1, \mathbf{z}_2, \ldots, \mathbf{z}_{N_\text{w}}\} \) for the granular-ball computing module. 
In this work, a three-layer Long Short-Term Memory (LSTM) network is used as the encoder. The final hidden states across layers are concatenated to form $\mathbf{z}_i$, enabling multilevel temporal dependency modeling. The encoder is modular and can be replaced with other models, such as Transformers \cite{Transformer}, for different time series characteristics.

\subsection{Granular-ball Vector Data Description}
After obtaining initial representations, granular-ball computing is introduced to adaptively model the distribution of latent time series data by constructing \emph{granular-ball vector descriptions} based on local data density and homogeneity, enabling flexible and effective normality modeling for one-class anomaly detection.

The granular-ball construction in GBOC follows the standard process described in Section \ref{sec:pre} but operates in the latent representation space, which is optimized during training. Unlike traditional granular-balls in the original data space \cite{GBC,W-GBC,GBG++}, traditional granular-balls are not directly suitable for normality modeling due to the following reasons: 
(i) They are not integrated into the representation learning process, limiting their ability to effectively guide feature learning.  
(ii) Some granular-balls may be of low quality or confidence, and their inclusion could degrade the overall model performance.  

To address these issues, we introduce two key components in our granular-ball representation optimization stage: (i) a pruning mechanism to remove low-quality granular-balls, and (ii) a joint learning strategy to align latent representations with high-quality granular-ball centers while maintaining temporal fidelity.

\subsection{Granular-ball Representation Optimization}\label{sec:loss}

\subsubsection{Elimination of Low-quality Granular-balls.}
To improve the compactness and reliability of the granular-ball representation, we propose a post-hoc pruning strategy that eliminates low-quality granular-balls characterized by abnormally large radii. Such granular-balls often correspond to overly coarse or poorly localized regions in the latent space, failing to capture fine-grained structural details and potentially introducing noise into the representation learning process.

The pruning process is governed by a dynamic threshold derived from the global distribution of granular-ball radii:
\begin{equation}
    r_{\text{th}} = \mu \cdot \max \left\{ \operatorname{median}(r), \operatorname{mean}(r) \right\},
\end{equation}
where \( r = \{ r_j \}_{j=1}^{|GB|} \) represents the set of radii for all granular-balls. 
$\mu=2$ empirically. 
Any granular-ball \( GB_j \) with radius \( r_j > r_{\text{th}} \) is deemed diffuse and subsequently removed. This radius-based pruning strategy effectively discards noisy or low-confidence regions while retaining structurally meaningful and compact granular-balls. These preserved granular-balls more accurately reflect the true distribution of normal data within the latent space.

\subsubsection{Nearest Granular-ball Center-Based Optimization.}
To ensure that learned representations are both semantically compact and temporally informative, we propose a joint optimization framework that combines a loss for preserving the consistency of granular-ball vector data descriptions with a loss for time series reconstruction.

The granular-ball loss functions as a geometric alignment mechanism, enforcing compact semantic clustering in the latent space by pulling each feature vector toward its corresponding granular-ball centroid. This aggregation mechanism enhances the discriminability of the latent space by clustering samples from the normal mode together, forming a compact feature space representation:
\begin{equation}
\mathcal{L}_{\text{gb}} = \frac{1}{N} \sum_{i=1}^{N} \left\| \mathbf{z}_i - \mathbf{c}_{s(i)} \right\|_2^2,
\label{eq:lossgb}
\end{equation}
where $s(i)$ denotes the index of the granular-ball assigned to the $i$-th sample, and $\mathbf{c}_{s(i)}$ is the corresponding center.

To complement this alignment objective and preserve temporal fidelity, we incorporate a \emph{reconstruction loss} that ensures the encoder-decoder pipeline retains essential information from the input. This constraint prevents the model from collapsing representations into overly compressed spaces that lack discriminative features:
\begin{equation}
\mathcal{L}_{\text{rec}} = \frac{1}{N} \sum_{i=1}^{N} \left\| \mathbf{x}_i - g_\phi(\mathbf{z}_i) \right\|_2^2,
\label{eq:lossrec}
\end{equation}
where \( g_\phi \) denotes a light-weight MLP parameterized by \( \phi \).

Finally, the overall training objective is defined as:
\begin{equation}
\mathcal{L} = \lambda \cdot \mathcal{L}_{\text{rec}} + (1-\lambda) \cdot \mathcal{L}_{\text{gb}},
\label{eq:obj}
\end{equation}
where \( \lambda \) is a trade-off coefficient (commonly set to 0.5) that balances the reconstruction fidelity and the strength of the structural alignment.

\begin{table*}[t!]
  \centering
  \scalebox{0.74}{
    \begin{tabular}{c|ccc|ccc|ccc|ccc|ccc|ccc}
    \toprule
    \multirow{2}[4]{*}{Models} & \multicolumn{3}{c|}{SMD Facility} & \multicolumn{3}{c|}{TAO Environment} & \multicolumn{3}{c|}{YAHOO Synthetic} & \multicolumn{3}{c|}{UCR Medical} & \multicolumn{3}{c|}{IOPS WebService} & \multicolumn{3}{c}{WSD WebService} \\
\cmidrule{2-19}          & VP    & VR    & AF    & VP    & VR    & AF    & VP    & VR    & AF    & VP    & VR    & AF    & VP    & VR    & AF   & VP    & VR    & AF\\
    \midrule
    PCA   & 0.482  & 0.821  & 0.666  & 0.951  & 0.966  & 0.004   & 0.024  & 0.370  & NaN   & 0.724  & 0.998  & \underline{0.996}   & 0.202  & 0.660  & 0.291  & 0.109  & 0.878  & 0.222 \\
    KNN   & 0.766  & 0.814  & 0.862  & 0.940  & 0.966  & 0.003  & 0.281  & 0.948  & NaN   & 0.856  & \textbf{1.000}  & 0.994  & 0.222  & 0.957  & 0.531  & 0.011  & 0.035  & NaN \\
    IForest    & 0.407  & 0.804  & 0.650  & 0.940  & 0.962  & 0.032  & 0.027  & 0.832  & 0.263  & 0.025  & 0.876  & NaN  & 0.335  & 0.917  & 0.478  & 0.024  & 0.614  & 0.173 \\
    MP    & 0.024  & 0.845  & 0.214  & 0.945  & 0.968  & NaN  & 0.347  & 0.955  & NaN  & 0.830  & 0.999  & 0.996  & 0.245  & 0.961  & 0.479  & 0.012  & 0.045  & NaN \\
    KShapeAD  & 0.019  & 0.799  & NaN  & 0.939  & 0.958  & NaN  & 0.048  & 0.383  & NaN  & \textbf{0.998}  & \textbf{1.000}  & 0.995  & 0.121  & 0.816  & 0.298  & 0.010  & 0.006  & NaN \\
    \midrule
    CNN   & 0.755  & 0.979  & 0.999  & \textbf{1.000}  & \textbf{1.000}  & 0.005  & 0.044  & 0.898  & 0.626  & 0.299  & 0.949  & 0.772  & 0.344  & 0.949  & 0.669  & \underline{0.609}  & 0.921  & \underline{0.981} \\
    LSTMAD  & 0.782  & 0.991  & \underline{0.999}  & \textbf{1.000}  & \textbf{1.000}  & NaN  & 0.087  & 0.959  & 0.413  & 0.151  & 0.981  & 0.691  & 0.386  & \underline{0.983}  & \underline{0.916}  & 0.249  & 0.817  & 0.834 \\
    TranAD  & 0.364  & 0.896  & 0.661  & 0.955  & 0.974  & NaN   & 0.091 & 0.967  & 0.415  & 0.015  & 0.788  & 0.590  & 0.307  & 0.958  & 0.745  & 0.122  & 0.721  & 0.711 \\
    USAD  & 0.369  & 0.821  & 0.669  & 0.950  & 0.966  & 0.006   & 0.123  & 0.928  & 0.157   & 0.020  & 0.854  & 0.634  & 0.314 & 0.949  & 0.663  & 0.093  & 0.807  & 0.497 \\
    TimesNet  & 0.680  & 0.961  & 0.997  & 0.932  & 0.978  & 0.015  & 0.577  & \underline{0.980}  & \underline{0.881}  & 0.023  & 0.875  & 0.653  & 0.184  & 0.906  & 0.811  & 0.354  & \underline{0.983}  & 0.805 \\
    DeepSVDD  & \underline{0.812}  & \underline{0.998}  & 0.866  & 0.945  & 0.961  & 0.116  & \underline{0.967}  & 0.896  & 0.071  & 0.996  & \textbf{1.000}  & 0.934  & 0.236  & 0.895  & 0.322  & 0.404  & 0.982  & 0.698 \\
    A.T.    & 0.017  & 0.494  & 0.401  & 0.935  & 0.950  & 0.005  & 0.098  & 0.864  & 0.651  & 0.135  & 0.755  & 0.655  & 0.124  & 0.891  & 0.685  & 0.354  & 0.393  & 0.511 \\
    THOC  & 0.272  & 0.891  & 0.663  & 0.938  & 0.962  & 0.042   & 0.048  & 0.698  & 0.275   & 0.513  & 0.979  & 0.836  & \underline{0.407}  & 0.906  & 0.331  & 0.025  & 0.555  & 0.092 \\
    MEMTO & 0.314  & 0.962  & 0.663  & 0.932  & 0.962  & 0.010   & 0.074  & 0.868  & 0.393  & 0.630  & 0.878  & 0.453  &0.180  & 0.912  & 0.743  &0.021  & 0.501  & NaN 
    \\
    \midrule
    \textbf{GBOC}  & \textbf{0.831}  & \textbf{0.999}  & \textbf{0.999}  & \underline{0.978}  & \underline{0.991}  & \textbf{0.219}  & \textbf{0.991} & \textbf{1.000} & \textbf{0.950} & \underline{0.996} & \underline{0.999} & \textbf{0.996}  & \textbf{0.604}  & \textbf{0.992}  & \textbf{0.948}  & \textbf{0.963}  & \textbf{0.998}  & \textbf{0.995} \\
    \bottomrule
    \end{tabular}
    }
    \caption{Univariate results on time series anomaly detection datasets (VP: VUS-PR, VR: VUS-ROC, AF: Affiliation-F1).}
    \label{tab:uni}
\end{table*}

\begin{table*}[t!]
  \centering
  \scalebox{0.74}{
    \begin{tabular}{c|ccc|ccc|ccc|ccc|ccc}
    \toprule
    \multirow{2}[4]{*}{Models} & \multicolumn{3}{c|}{LTDB Medical} & \multicolumn{3}{c|}{SVDB Medical} & \multicolumn{3}{c|}{TAO Environment} & \multicolumn{3}{c|}{SMD Facility} & \multicolumn{3}{c}{SMAP Sensor} \\
\cmidrule{2-16}          & VP    & VR    & AF    & VP    & VR    & AF    & VP    & VR    & AF    & VP    & VR    & AF    & VP    & VR    & AF \\
    \midrule
    PCA   & 0.131  & 0.310  & 0.076  & \textbf{0.416}  & \textbf{0.655}  & 0.517   & 0.154  & 0.389  & NaN   & 0.182  & 0.777  & 0.182  & 0.019  & 0.579  & 0.773 \\
    KNN   & 0.149  & 0.490  & 0.124  & 0.109  & 0.511  & 0.442  & 0.141  & 0.012  & 0.048   & 0.182  & 0.734  & \underline{0.864}  & 0.014  & 0.499  & 0.650  \\
    IForest  & \underline{0.208}  & \underline{0.551}  & 0.510  & 0.239  & \underline{0.640}  & \underline{0.859}  & 0.146  & 0.362  & 0.057   & 0.118  & 0.640  & 0.322  & 0.014  & 0.561  & 0.628 \\
    \midrule
    CNN   & 0.140  & 0.482  & 0.456  & 0.113  & 0.555  & 0.664  & \underline{0.996}  & \underline{0.999}  & NaN  & 0.202  & 0.750  & 0.180  & 0.019  & 0.574  & 0.602 \\
    LSTMAD    & 0.135  & 0.457  & 0.408  & 0.109  & 0.544  & 0.665  & 0.996  & 0.999  & NaN  & 0.185  & 0.695  & 0.180  & 0.018  & 0.543  & 0.602 \\
    TranAD  & 0.136  & 0.462  & 0.424  & 0.104  & 0.522  & 0.665  & 0.144  & 0.364  & 0.002  & 0.168  & 0.573  & 0.181  & 0.017  & 0.469  & 0.601 \\
    USAD  & 0.137  & 0.460  & 0.379  & \underline{0.384}  & 0.643  & 0.382  & 0.255  & 0.629  & 0.014  & 0.158  & 0.730  & 0.182  & 0.017  & 0.574  & 0.782 \\
    TimesNet  & 0.112  & 0.385  & 0.550  & 0.111  & 0.564  & 0.657  & 0.165  & 0.433  & \underline{0.179}  & \underline{0.309}  & \underline{0.831}  & 0.182  & 0.017  & 0.551  & 0.597 \\
    A.T.  & 0.147  & 0.497  & \underline{0.559}  & 0.098  & 0.488  & 0.641  & 0.141  & 0.488  & 0.055  & 0.036  & 0.332  & 0.651  & 0.014  & 0.496  & 0.180 \\
    DeepSVDD  & 0.198  & 0.469  & 0.169  & 0.129  & 0.562  & 0.049  & 0.160  & 0.414  & 0.030  & 0.188  & 0.805  & 0.181  & 0.023  & 0.646  & 0.536 \\
    THOC  & 0.144  & 0.369  & 0.142  & 0.176  & 0.565  & 0.177   & 0.142  & 0.338  & 0.021  & 0.052  & 0.508  & 0.303  & \underline{0.310}  & \underline{0.681}  & \underline{0.791} \\
    MEMTO  & 0.147  & 0.450  & 0.276  & 0.114  & 0.536  & 0.377   & 0.151  & 0.374  & 0.013   & 0.104  & 0.685  & 0.178  & 0.024  & 0.652  & 0.739 \\
    \midrule
    \textbf{GBOC}  & \textbf{0.241}  & \textbf{0.564}  & \textbf{0.747}  & 0.248  & 0.572  & \textbf{0.877}  & \textbf{0.996} & \textbf{0.999} & \textbf{0.182} & \textbf{0.369} & \textbf{0.851} & \textbf{0.898}  & \textbf{0.338}  & \textbf{0.723}  & \textbf{0.833} \\
    \bottomrule
    \end{tabular}
    }
    \caption{Multivariate results on time series anomaly detection datasets (VP: VUS-PR, VR: VUS-ROC, AF: Affiliation-F1).}
    \label{tab:multi}
\end{table*}

\subsection{Anomaly Inference}
During inference, each test instance \( \mathbf{w} \) is encoded into a representation \( \mathbf{z} = f_\theta(\mathbf{w}) \). The degree of deviation from the normal region is quantified by computing the Euclidean distance between \( \mathbf{z} \) and the nearest granular-ball center:
\begin{equation}
\text{Score}(\mathbf{z}) = \min_{\mathbf{c} \in \mathcal{C}} \left\| \mathbf{z} - \mathbf{c} \right\|_2,
\end{equation}
where \( \mathcal{C} \) denotes the set of retained granular-ball centers. A higher anomaly score indicates a greater deviation from the known normal patterns. Samples with scores exceeding a predefined threshold are classified as anomalies. 
To determine anomaly labels in an unsupervised manner, we adopt a statistical thresholding strategy. Specifically, we apply the empirical \( 3\sigma \) rule \cite{3sigma}, whereby a test sample is flagged as anomalous if its score exceeds three standard deviations above the mean anomaly score observed on the evaluation set. This simple yet effective strategy provides a principle-based decision boundaries without relying on labeled anomalies \cite{MSL, 3sigma2}.

\noindent \textbf{Discussions.}
Related clustering-based methods rely on rigid boundaries, predefined cluster structures, or simplistic single-hypersphere assumptions, making them less effective for dynamic, continuous, or multimodal time series. Even multiscale approaches like THOC~\cite{THOC}, which employ hierarchical clustering for temporal features, are limited by fixed structures, reducing adaptability to density and temporal variations.

In contrast, the proposed GBOC dynamically forms granular-ball structures that adaptively capture both local density and global continuity without requiring predefined parameters or rigid boundaries. This ensures robust and effective anomaly detection in complex and noisy real-world time series distributions.

\section{Experiment}
In this section, we analyze the experimental results of the proposed method on seven widely used time series anomaly detection datasets
and compare it with several state-of-the-art methods to
demonstrate its effectiveness.

\subsection{Experimental Settings}

\subsubsection{Datasets.}
We include both univariate and multivariate time series anomaly detection benchmark datasets. 
These datasets span a broad range of real-world domains, including industrial systems (SMD \cite{OmniAnomaly}), web services (IOPS \cite{TSBAD}, WSD \cite{WSD}), healthcare (UCR \cite{UCR}, LTDB \cite{LTDB}, SVDB \cite{SVDB}), environmental sensing (TAO \cite{TSBAD}, SMAP \cite{MSL}, MSL \cite{MSL}), and synthetic monitoring (YAHOO \cite{YAHOO}). This diversity ensures that our evaluation captures both point and range-based anomalies, as well as varied temporal dynamics (e.g., high nonlinearity with noise, complex temporal variations, and distribution shifts) and different data characteristics. For all datasets, we follow the official train/tune/test splits provided by \cite{TSBAD} to ensure fairness. 
See Appendix \ref{sec:apendixdata} for more detailed descriptions of the datasets.

\subsubsection{Compared Methods.}
We compare our method with 14 baselines, which are divided into two main groups: non-deep learning methods and deep learning methods. (i) The former baselines include 
PCA \cite{SubPCA}, KNN \cite{SubKNN}, 
IForest \cite{isolationforest}, MatrixProfile \cite{Matrixprofile} and KShapeAD \cite{KshapeAD2}. (ii) The latter includes CNN \cite{CNN}, LSTMAD \cite{LSTMAD}, TranAD \cite{Tranad}, USAD \cite{USAD}, TimesNet \cite{TimesNet} and AnomalyTransformer (A.T.) \cite{AnomalyTrans}. Moreover, we also include three competitive models such as DeepSVDD \cite{deepsvdd}, THOC \cite{THOC}, and MEMTO \cite{MEMTO}. See Appendix \ref{sec:apendixbas} for more detailed descriptions of the baselines.

\subsubsection{Implementation Details.}
Following the protocols in \cite{TSBAD}, random seed is set for 2024, we tune the window size and the number of LSTM encoder layers for the proposed GBOC model based on validation performance. The window size is selected from ${2, 5, 10, 50}$ based on the temporal resolution and dataset length, while the number of LSTM layers are set between 1 and 3, depending on input complexity. The hidden layer dimension is fixed at 32, and the loss weight $\lambda$ is set to 0.5 by default. These settings remain consistent across datasets unless stated otherwise. All models are trained using the Adam optimizer with a learning rate of $10^{-4}$ and a batch size of 32. The hyperparameters of the baselines are shown in Appendix \ref{sec:baspara}. All experiments were conducted on a workstation equipped with an NVIDIA RTX 4090 GPU and 128 GB of RAM, using Python 3.10 and PyTorch 1.13.

\subsubsection{Evaluation Metrics.}
We evaluate anomaly detection models using three metrics: VUS-PR \cite{VUS-ROC(PR)}, VUS-ROC \cite{VUS-ROC(PR)}, and Affiliation-F1 \cite{AF1}, reflecting ranking quality, temporal robustness, and range-based alignment. VUS-PR, based on the precision-recall manifold, is more suitable for rare-event scenarios than AUC-PR \cite{AUC-PR}. VUS-ROC introduces a tolerance window $\Delta$ to account for detection delays, treating nearby predictions as correct. Affiliation-F1 evaluates the alignment between predicted anomalies and ground truth intervals using a Gaussian distance kernel. However, some baselines achieve high VUS-PR or VUS-ROC scores but yield NaN in Affiliation-F1 when predictions fail to overlap with ground truth, as shown in Figure \ref{fig:AffNaN1}, we also provide a detailed analysis in Appendix \ref{sec:apendixf1nan}. Thus, these metrics complement each other to provide full insight into model performance.
\begin{figure}[t!]
    \centering  \includegraphics[width=0.95\linewidth]{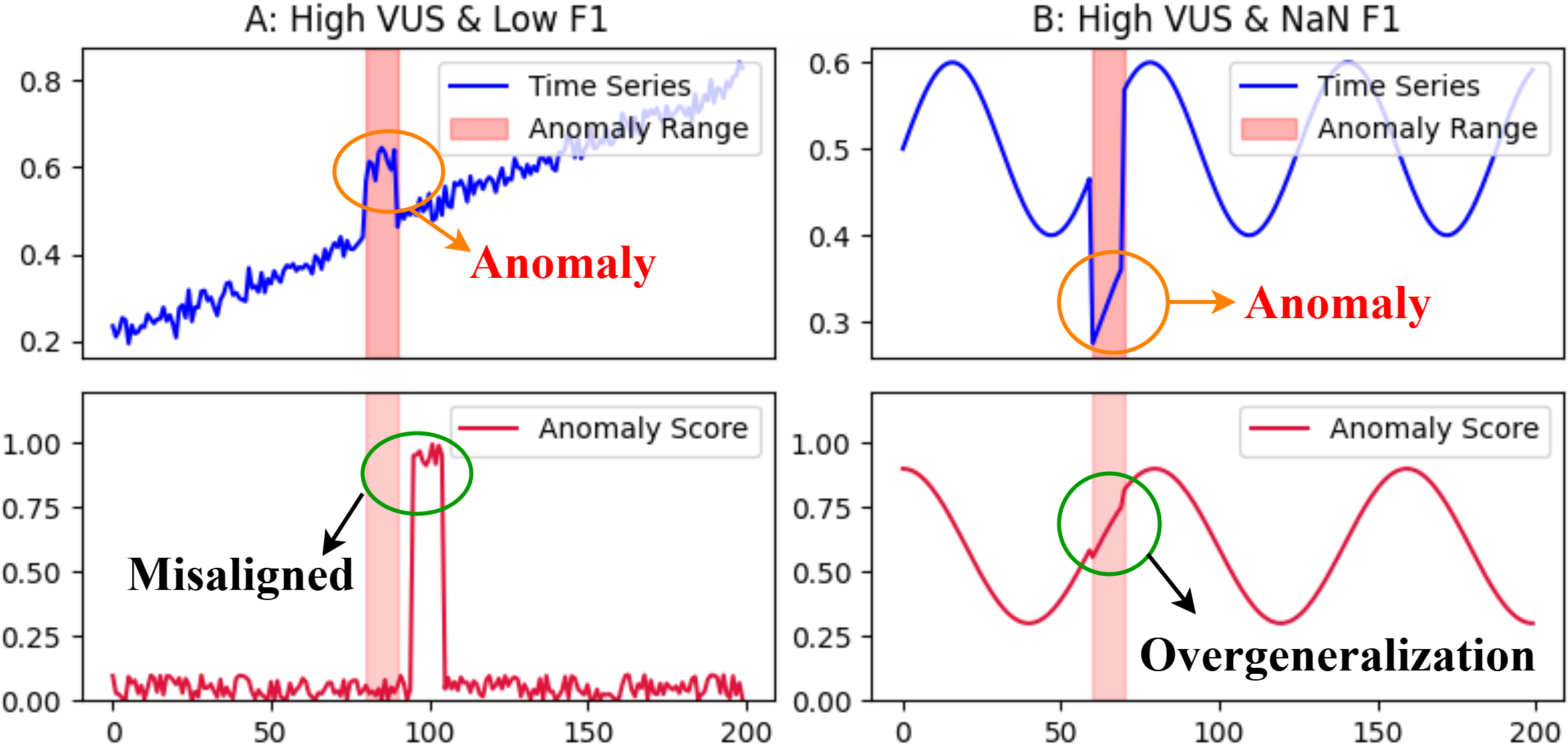}
    \caption{(a) shows high VUS but low Affiliation-F1 due to misaligned anomaly scores. (b) shows high VUS but NaN F1 as no predicted scores overlap with anomalies.}
    \label{fig:AffNaN1}
\end{figure}

\subsection{Main Results}

\subsubsection{Univariate Detection.}
As shown in Table \ref{tab:uni}, GBOC consistently achieves strong performance on univariate time series anomaly detection tasks, securing the highest overall average score. It excels particularly on datasets such as SMD, YAHOO, IOPS, and WSD, while maintaining competitive accuracy across other benchmarks. These results highlight GBOC's ability to handle diverse univariate anomaly detection challenges effectively.

\subsubsection{Multivariate Detection.}
In Table \ref{tab:multi}, the proposed GBOC demonstrates exceptional performance in multivariate anomaly detection, outperforming several state-of-the-art baselines, including TimesNet, A.T., DeepSVDD, THOC, and MEMTO. Its robust results across various datasets reinforce its versatility and effectiveness in identifying anomalies in complex multivariate settings.


\subsection{Robust Detection under Drift and Noise}
This section examines four scenarios with different temporal dynamics: Type I (Clean, no drift or noise), Type II (Drift-only), Type III (Noise-only), and Type IV (Drift and noise). Table \ref{tab:case-study-results} summarizes the results:
\begin{table}[t!]
  \centering
  \newcolumntype{C}{>
  {\centering\arraybackslash}p{0.08\textwidth}}
  \scalebox{0.81}{
    \begin{tabular}{cCCCc}
    \toprule
    \multirow{1}[1]{*}{Methods} & \multicolumn{1}{c}{I: clean} & \multicolumn{1}{c}{II: drfit} & \multicolumn{1}{c}{III: noise} & \multicolumn{1}{c}{IV: drift + noise} \\
    \midrule
    PCA   & 0.955  & 0.005   &  0.331   & 0.083 \\
    KNN   & 0.991  & 0.586 & 0.591   & 0.442  \\
    IForest  &  0.217 & 0.029 &  0.082  & 0.065 \\
    MP    & 0.995  & 0.594  & 0.640  & 0.450  \\
    KShapeAD  & \underline{1.000}  & \textbf{0.982}  & 0.802  & 0.624  \\
    \midrule
    CNN   & 0.502  & 0.013  & 0.230  & 0.069  \\
    LSTMAD  & 0.503  & 0.005  &  0.280  & 0.143  \\
    TranAD  & 0.455  & 0.005   &  0.101  & 0.173  \\
    USAD  & 0.824  & 0.004   & 0.061   & 0.103 \\
    TimesNet  & 0.747  &  0.819   & 0.037  & 0.042  \\
    A.T.    & 0.479  & 0.011  & 0.532   & 0.474  \\
    DeepSVDD    & 0.824  & 0.153  & \underline{0.833}   & \underline{0.893}  \\
    THOC  & 0.569  & 0.006   & 0.088   & 0.075 \\
    MEMTO & 0.782  & 0.028   & 0.121   & 0.031 \\
    \midrule
    \textbf{GBOC}   & \textbf{1.000}  & \underline{0.977}  & \textbf{0.952} & \textbf{0.921} \\
    \bottomrule
    \end{tabular}
     }
    \caption{{VUS-PR results on four real-world anomaly types datasets from MSL (I), SMAP (II), and YAHOO (III, IV).}}
    \label{tab:case-study-results}
\end{table}

\textbf{Type I (Clean)}: Nearly all models perform well, with GBOC and KShapeAD achieving near-perfect accuracy, reflecting their ability to handle well-structured data.

\textbf{Type II (Drift-only)}: Models that rely on stationary assumptions, such as PCA, LSTM, and CNN, suffer significant performance declines due to their inability to adapt to distributional shifts. In contrast, GBOC and KShapeAD demonstrate strong adaptability, maintaining high performance despite the drift.

\textbf{Type III (Noise-only)}: GBOC emerges as the most robust, leveraging its granular-ball-based modeling to isolate dense, high-quality regions and minimize false positives. This gives GBOC a significant edge over others. KShapeAD shows reasonable robustness but still experiences a marked performance drop due to its sensitivity to noise.

\textbf{Type IV (Drift and noise)}: This is the most challenging scenario, where all models face severe degradation. However, GBOC remains the top performer, showcasing its exceptional ability to generalize under compounded corruption. DeepSVDD demonstrates commendable resilience, but most other baselines, including traditional clustering and KNN methods, struggle heavily.

Overall, GBOC demonstrates consistent top performance across scenarios, particularly in noisy or nonstationary environments, where traditional methods like KShapeAD degrade significantly.

\subsection{Ablation Studies}
In this section, we perform ablation studies to analyze:  
(i) the impact of granular-ball data description and  
(ii) the roles of losses $\mathcal{L}_{\text{rec}}$ and $\mathcal{L}_{\text{gb}}$ in Equation (\ref{eq:obj}). 
More ablation study results are provided in Appendix \ref{sec:apendixabl}.

\subsubsection{Granular-ball Data Description.}   
Two GBOC variants were tested:  
(1) \textit{w/o granular-ball computing}: Replaces granular-ball construction with K-Means.  
(2) \textit{w/o granular-ball pruning}: Retains all granular-balls without pruning.   
As shown in Table \ref{tab: Loss-gb}, removing pruning degrades the performance by retaining noisy regions. Using K-Means results in greater drops due to its fixed clusters, making it less adaptable to varying data densities. In contrast, GBOC dynamically adjusts to local densities, enabling superior generalization to complex and irregular data distributions.
\begin{table}[t!]
  \centering
  \scalebox{0.79}{
  \begin{tabular}{cc|ccccccc}
    \toprule
    $\mathcal{\text{GBC}}$ & $\mathcal{\text{Pruning}}$ & SMD    & IOPS   & UCR   & YAHOO  & TAO   & WSD  \\
    \midrule
    $\times$ & $\times$ & 0.755  & 0.554  & 0.921  & 0.823  & 0.948 & 0.911 \\
    $\checkmark$     & $\times$ & 0.781  & 0.566  & 0.972  & 0.795  & 0.959 & 0.885 \\
    $\checkmark$ & $\checkmark$ & \textbf{0.831}  & \textbf{0.604}  & \textbf{0.996}  & \textbf{0.991}  & \textbf{0.978} & \textbf{0.963} \\
    \bottomrule
  \end{tabular}
  }
  \caption{Effects of \textit{w/o granular-ball computing} and \textit{w/o pruning} regarding VUS-PR.}
  \label{tab: Loss-gb} 
\end{table}

\begin{table}[t!]
  \centering
  \scalebox{0.8}{
  \begin{tabular}{cc|ccccccc}
    \toprule
    $\mathcal{L}_{\text{rec}}$ & $\mathcal{L}_{\text{gb}}$ & SMD    & IOPS   & UCR   & YAHOO  & TAO   & WSD  \\
    \midrule
    $\checkmark$ & $\times$ & 0.780  & 0.545  & 0.955  & 0.869  & 0.952 & 0.936 \\
    $\times$     & $\checkmark$ & 0.715  & 0.506  & 0.939  & 0.701  & 0.964 & 0.945 \\
    $\checkmark$ & $\checkmark$ & \textbf{0.831}  & \textbf{0.604}  & \textbf{0.996}  & \textbf{0.991}  & \textbf{0.978} & \textbf{0.963} \\
    \bottomrule
  \end{tabular}
  }
  \caption{Effects of $\mathcal{L}_{\text{rec}}$ and $\mathcal{L}_{\text{gb}}$ regarding VUS-PR .} 
  \label{tab: Loss-VP} 
\end{table}

\subsubsection{Effects of $\mathcal{L}_{\text{rec}}$ and $\mathcal{L}_{\text{gb}}$.} 
We evaluated:  
(1) \textit{Reconstruction loss $\mathcal{L}_{\text{rec}}$}: Preserves input structure in latent representations.  
(2) \textit{Granular-ball optimization loss $\mathcal{L}_{\text{gb}}$}: Aligns embeddings with granular-ball centers. 
As shown in Table \ref{tab: Loss-VP}, removing $\mathcal{L}_{\text{gb}}$ leads to a drop in VUS-PR values due to disorganized embeddings and reduced discriminative ability, while excluding $\mathcal{L}_{\text{rec}}$ compromises semantic coherence. These effects are particularly evident on the noisy YAHOO dataset, underscoring the importance of both losses.

\subsection{Hyperparameter Sensitivity Analysis}
In this section, sensitivity analysis regarding VUS-PR on three hyperparameters are performed. They are (i) number of number of LSTM encoder layers, (ii) number of sliding window and (iii) loss weight $\lambda$ in Equation \ref{eq:obj}. More results are provided in Appendix \ref{sec:apendixhyper}.

\begin{figure}[h!]
    \centering  \includegraphics[width=1\linewidth]{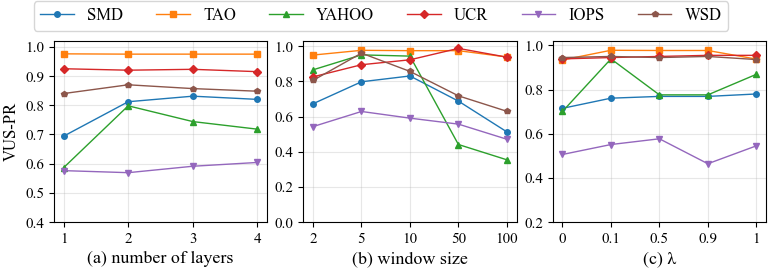}
    \caption{{Sensitivity analysis of GBOC on VUS-PR to the number of LSTM encoder layers, input window size and loss weight $\lambda$ across multiple datasets.}}
    \label{fig:hyper}
\end{figure}

As shown in Figure \ref{fig:hyper} (left), the number of layers is a relatively insensitive hyperparameter. In most cases, a two-layer LSTM network achieves satisfactory performance.

In Figure \ref{fig:hyper} (middle), the window size  demonstrates varying sensitivity across different datasets. For example, a small window size performs better on YAHOO dataset, while a window size of 50 yields better results for UCR dataset. 

By varying $\lambda$ from 0 to 1 in Figure \ref{fig:hyper} (right), we observe that selecting an appropriate value for $\lambda$ achieves better results compared to $\lambda = 0$ or $\lambda = 1$, demonstrating the effectiveness of this trade-off parameter.

\begin{figure}[t!]
    \centering  
    \includegraphics[width=0.47\textwidth, height=1.7in]{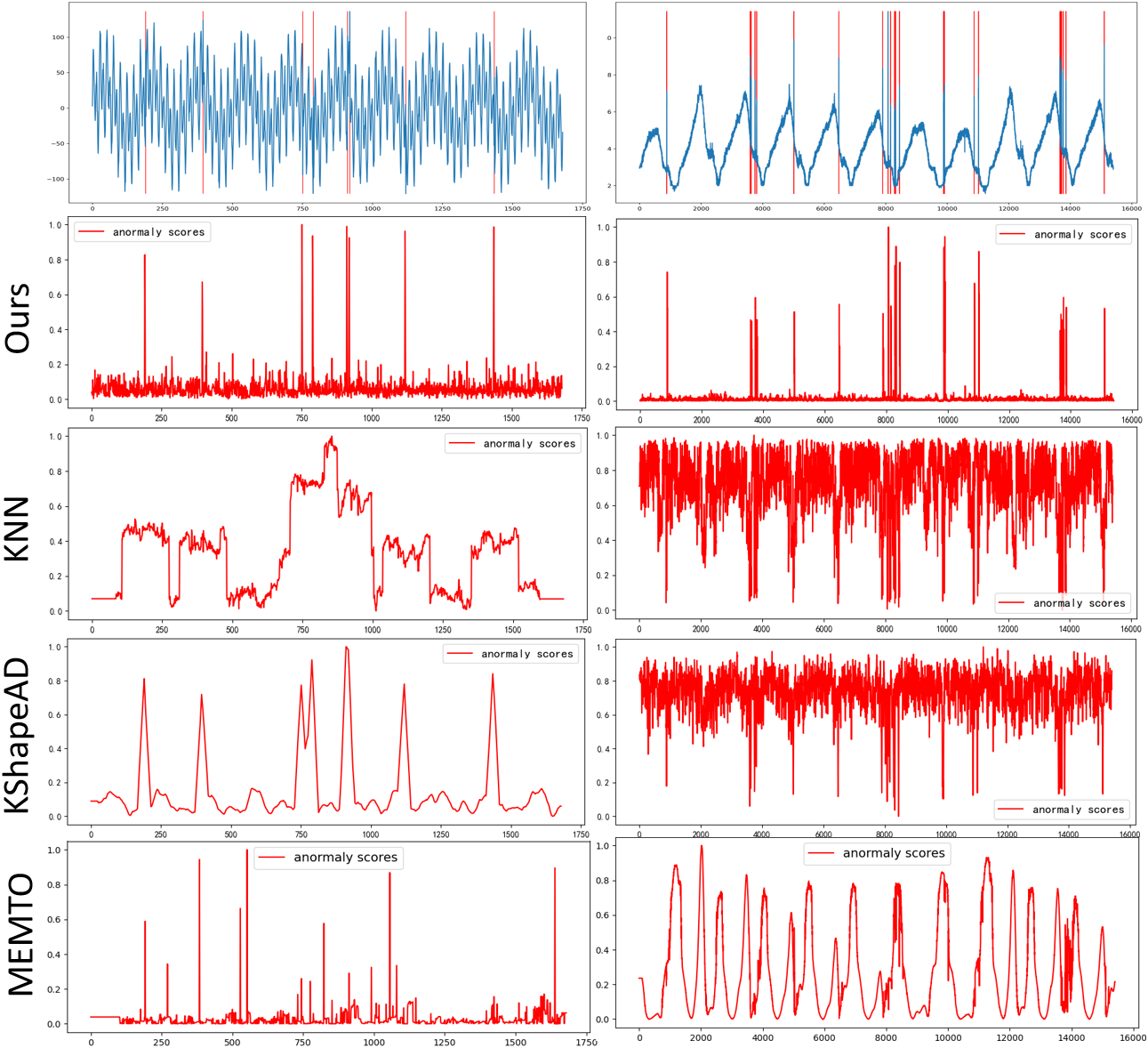}
    \caption{{Visualization of anomaly scores on two representative datasets from YAHOO and WSD, comparing GBOC with KNN, KShapeAD, and MEMTO. Ground-truth anomalies are shaded in red.}}
    \label{fig:Visual}
\end{figure}

\subsection{Visualization Analysis}
To assess GBOC's robustness and interpretability, we visualize its anomaly scores on YAHOO and WSD, and compare them with three baselines: KNN, KShapeAD, and MEMTO. As shown in Figure \ref{fig:Visual}, GBOC produces smooth, well-localized scores that align closely with the ground truth. In contrast, KShapeAD and KNN often yield fluctuating scores in normal regions with noise or periodic variations, causing false positives. MEMTO, while more stable, tends to blur anomaly boundaries and mislabel extended normal regions. GBOC suppresses spurious spikes caused by noise, thanks to its compact, high-density granular-ball memory, which filters noise at the representation level. These results confirm GBOC's quantitative advantages and highlight its interpretability and consistency across diverse scenarios. 
See Appendix \ref{sec:apendixas} for more visualization results of anomaly scores.

\section{Conclusion}
In this work, we proposed GBOC, a novel one-class network for time series anomaly detection based on adaptive granular-ball vector data description (GVDD). By organizing latent representations into compact, high-density granular-balls, GBOC captures local structure effectively and provides a flexible, interpretable one-class data description. Additionally, we introduced a granular-ball representation optimization mechanism.  Experiments on diverse benchmarks show that GBOC achieves competitive or superior performance, particularly excelling under noisy and nonstationary conditions. Ablation studies validated the contributions of its components, while case studies highlight its adaptability across anomaly scenarios. In conclusion, GBOC achieves strong detection performance with flexible distribution description, offering a promising direction for unsupervised time series analysis.

\section{Acknowledgments}
{
This work was supported in part by the National Natural
Science Foundation of China under Grant Nos. 62221005,
62450043, 62222601, 62176033 and 62576056.
}

\bibliography{aaai2026}

\appendix
\onecolumn
\section{Related Work}\label{sec:related}
\subsection{Time Series Anomaly Detection}
Time series anomaly detection (TSAD) has been extensively explored across domains such as industrial monitoring, finance, and healthcare. Traditional techniques primarily include statistical and distance-based approaches, such as Local Outlier Factor (LOF)\cite{LOF}, Principal Component Analysis (PCA)\cite{SubPCA}, K-Nearest Neighbors (KNN)\cite{SubKNN}, and MatrixProfile\cite{Matrixprofile}. These methods typically assume that normal patterns follow a stable distribution or lie within a low-dimensional subspace, enabling anomaly detection as deviations from this structure. While effective in stationary or low-noise environments, their performance degrades significantly on complex, high-dimensional, or nonstationary time series. Nearest-neighbor approaches, for example, struggle to capture temporal dynamics and are prone to misclassifying group anomalies due to their reliance on local density estimation. Such challenges are further exacerbated by distribution shifts and irregular temporal behaviors, which are common in real-world data.

The rise of deep learning has introduced more expressive models for TSAD. Reconstruction-based methods, such as USAD~\cite{USAD} and OmniAnomaly~\cite{OmniAnomaly}, utilize autoencoders to model normal sequences and flag anomalies via reconstruction errors. Prediction-based approaches, including LSTMAD~\cite{LSTMAD} and TranAD~\cite{Tranad}, detect anomalies by comparing predicted values to actual observations. These models excel at capturing nonlinear dependencies but often falter in noisy or nonstationary settings where subtle temporal cues are crucial.

Clustering-based strategies offer another unsupervised direction, identifying anomalies as samples residing outside dense data clusters. Representative methods include KMeansAD~\cite{KMeansAD} and KShapeAD~\cite{KshapeAD1, KshapeAD2, SAND}. Similarly, one-class classification models, such as OC-SVM~\cite{OCSVM} and SVDD~\cite{SVDD}, define a hypersphere to enclose normal points and treat those outside as anomalies. Memory-based frameworks extend this concept by storing prototype representations and scoring new samples by their proximity to these stored entries. Despite their conceptual appeal, these methods face several limitations. Clustering-based models require predefined cluster numbers and often assume discrete modes of normal behavior, which may not hold in dynamic sequences. One-class classifiers impose rigid geometric constraints, while memory-based models rely on the completeness and quality of stored prototypes, which may not generalize to unseen patterns.

To improve adaptability and generalization, recent works have introduced more flexible architectures. THOC~\cite{THOC}, for instance, extracts multiscale temporal features through hierarchical clustering and recurrent layers. MEMTO~\cite{MEMTO} incorporates memory-aware Transformers to better capture semantic and temporal dependencies. However, these models often rely on static memory layouts or handcrafted structures, limiting their ability to adapt to local density variations or noise.

\subsection{Granular-ball Computing} \label{sec:GBC}
Granular-ball Computing (GBC) is a flexible, geometry-aware framework originally developed for clustering and structure learning. It has demonstrated effectiveness in a range of tasks, including fast K-Means acceleration~\cite{xia2020fast}, improved density-based clustering~\cite{cheng2023fast}, and center-consistent adaptive structure learning~\cite{GBCT}.

Recently, GBC has been adopted in domains such as point cloud registration~\cite{hu2025gricp}, {granular-ball induced multiple kernel k-means clustering~\cite{GBMKKM}, granular-ball based open intent classification~\cite{MOGB}}, and multi-view contrastive learning~\cite{su2025multi}, underscoring its versatility as a general-purpose granular representation framework. Its capacity to adaptively model complex geometric structures without strong distributional assumptions makes it particularly promising for anomaly detection in nonstationary and noisy time series. The core procedure for granular-ball construction is described in Algorithm~\ref{alg:granular-ball}. 

While traditional time series anomaly detection (TSAD) methods, including reconstruction-based, prediction-based, clustering-based, and one-class classification models offer useful perspectives, they often suffer from critical limitations such as rigid distributional assumptions, sensitivity to hyperparameter tuning (e.g., number of clusters), reliance on fixed memory structures, and poor adaptability to local density variations or temporal nonstationarity. These issues are particularly pronounced in real-world time series, where data distributions are often dynamic and corrupted with noise or drift.

Granular-ball Computing (GBC) addresses many of these challenges by adaptively constructing data-dependent representations without strong parametric assumptions. Its hierarchical, density-aware partitioning enables flexible modeling of complex structures, while its ability to identify and prune noisy or diffuse regions makes it well-suited for learning robust representations in nonstationary environments.

Motivated by these complementary insights, we propose the Granular-ball One-Class Network (GBOC), a novel TSAD framework that leverages GBC to dynamically organize latent features into compact, high-density granular-balls, enabling robust, interpretable, and geometry-aware anomaly detection across diverse time series scenarios.

\begin{algorithm}[h]
\caption{Granular-ball generation process.}
\label{alg:granular-ball}
\begin{algorithmic}[1]
\STATE\textbf{Input} points $\{\rvx_t\}_{t=1}^n$; Minimum support threshold for granular-balls $s_{\min}$.\\
\STATE\textbf{Output} set of granular-balls $GB = \{ GB_j \}_{j=1}^m$. 
\STATE Initialize $k_0 \leftarrow \lfloor \sqrt{n} \rfloor$; \STATE Initialize $GB = \{ GB_j \}_{j=1}^{k_0}$ using $k_0$-Means;
\WHILE{True}
    \STATE $m \leftarrow |GB|$; 
    \FOR{each $GB_j \in GB$ with $|GB_j| > s_{\min}$}
        \STATE Split $GB_j$ into $GB_j^{(1)}$, $GB_j^{(2)}$ via 2-Means; 
        \STATE Compute $DM$, $DM_w$ via Equation (\ref{eq:DM}) and Equation~(\ref{eq:DMW}); 
        \IF{$DM_w < DM$}
            \STATE Replace $GB_j$ with $GB_j^{(1)}$, $GB_j^{(2)}$ in the set $GB$; 
        \ENDIF
    \ENDFOR
    \IF{$m$ equals to $|GB|$} \STATE \textbf{break} \ENDIF
\ENDWHILE
\STATE \textbf{return} $GB$
\end{algorithmic}
\end{algorithm}

\newpage
\section{More Disscussion}
\subsection{GBOC vs. Nearest-neighbor-based methods:}
Nearest-neighbor-based methods are a widely used class of anomaly detection approaches, which measure the distance between a test sample and its closest neighbors in the training set to assess normality. While conceptually simple and effective in certain settings, these methods often rely on the assumption that training neighbors are all normal and uniformly trustworthy. In real-world scenarios especially with noisy or partially contaminated training data, this assumption may introduce significant risk, as outliers or ambiguous samples can distort the decision boundary.

In contrast, GBOC builds its representation upon locally dense and homogeneous regions identified through granular-ball construction. By focusing on high-density areas, GBOC mitigates the influence of outliers and avoids over-reliance on potentially noisy individual neighbors. This density-aware formulation offers a more robust and principled estimation of normality, particularly for time series data with complex temporal dynamics and continuous latent transitions.

\begin{figure*}[h]
    \centering  \includegraphics[width=0.92\textwidth]{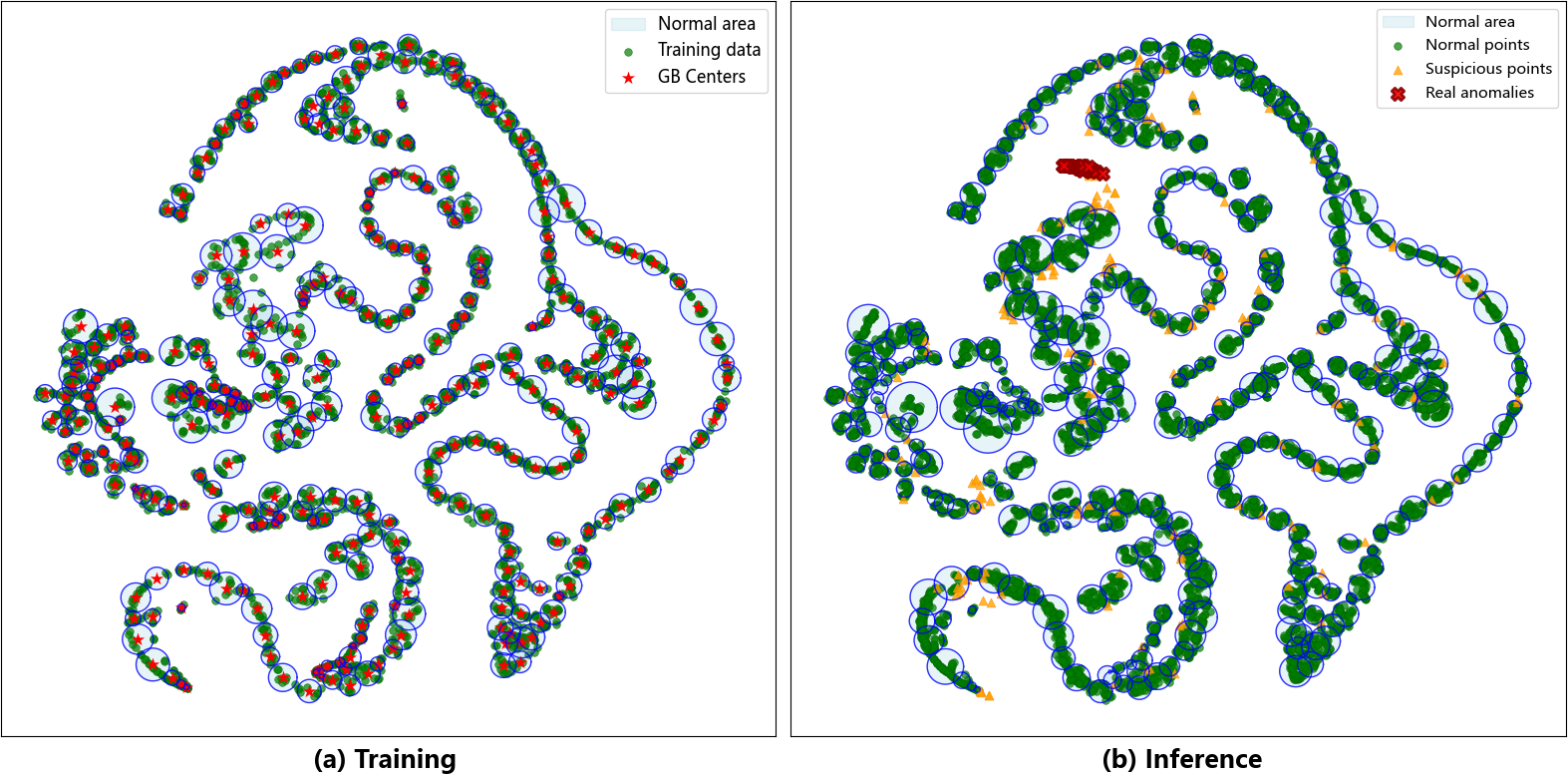}
    \caption{{Latent space visualization of GBOC on a time series dataset from UCR dataset using t-SNE projection. The left plot shows training samples (green) and learned granular-ball centers (red stars), which form compact, high-density regions representing normal patterns. The right plot depicts test results: normal samples lie within granular-ball regions, suspicious anomalies (orange triangles) fall near the boundary, and true anomalies (red crosses) are distant outliers. }}
    \label{fig:Visual_STNE}
\end{figure*}

Figure~\ref{fig:Visual_STNE} demonstrates how GBOC constructs a robust granular-ball-based representation by effectively excluding the influence of suspicious anomalies (i.e., false negative samples) during the modeling process. It further shows that GBOC captures the continuous latent structure of time series data and identifies anomalies as deviations from semantically coherent high-density regions.

\subsection{GBOC vs. Clustering-based models:}
Moreover, clustering-based models (also the memory-based and one-class models) typically assume that normal data reside in discrete, well-separated regions of the latent space, whether stored as fixed memory slots, grouped into explicit clusters, or enclosed within a global hypersphere. However, as shown in Figure \ref{fig:Visual_STNE}, time series data encoded via sliding windows naturally exhibit smooth and continuous transitions in latent space. This structural continuity challenges rigid partitioning strategies, since normal patterns rarely conform to clearly separated clusters or fixed geometric boundaries.

GBOC is explicitly designed to accommodate such continuity by organizing latent representations into adaptive, high-density granular-balls. Rather than enforcing hard class assignments or global constraints, GBOC models normality as a collection of semantically coherent regions that emerge organically from the data distribution. This design enables the model to flexibly represent fine-grained normal patterns and maintain robustness under noise, temporal variation, or distributional shifts, conditions under which traditional approaches often fail to generalize effectively.

\subsection{GBOC vs. Class-wise MOGB \cite{MOGB}:}
Although both MOGB and GBOC are grounded in granular-ball computing, they solve different problems and thus assign different roles to granular-balls. MOGB targets open intent classification and uses multi-granularity granular-balls as decision boundaries to refine \textit{class-wise} regions and to balance empirical risk and open-space risk under labeled intents.

By contrast, GBOC focuses on time series anomaly detection, where \textit{no explicit class structure exists}. It builds the granular-ball vector data description to partition the latent space into adaptive, high-density regions that describe normal patterns, and then prunes diffuse balls to improve robustness under drift and noise. In this sense, MOGB is a granular decision framework for open classification, while GBOC is a granular representation framework for density-aware one-class modeling.

\newpage
\section{Detailed Experimental Setup}
\subsection{Datasets} \label{sec:apendixdata}
To ensure broad coverage and evaluate generalizability, we select a diverse set of benchmark datasets from the TSB-AD \cite{TSBAD} benchmark suite. These datasets span multiple domains including industrial systems (e.g., SMD), healthcare (e.g., LTDB, SVDB), web services (e.g., IOPS, WSD), and synthetic benchmarks (e.g., YAHOO). For each dataset, we select representative subsets covering various types of anomalies such as point anomalies, range anomalies, noise-corrupted events, and distributional shifts. The datasets include both univariate and multivariate time series, allowing us to evaluate our method across different modalities. The detailed dataset is described as follows:
\begin{itemize}
    \item \textbf{SMD} \cite{OmniAnomaly}: a 5-week-long dataset collected from a large Internet company, which contains 3 groups of entities from 28 different machines.
    \item \textbf{TAO} \cite{TSBAD}: contains 575, 648 records with 3 attributes which are collected from the Tropical Atmosphere Ocean project.
    \item \textbf{YAHOO} \cite{YAHOO}: a dataset published by YAHOO labs consisting of real and synthetic time series based on the real production traffic to some of the YAHOO production systems.
    \item \textbf{UCR} \cite{UCR}: a collection of univariate time series of multiple domains including air temperature, arterial blood pressure, ABP, astronomy, EPG, ECG, gait, power demand, respiration, walking accelerator. Most of the anomalies are introduced artificially.
    \item \textbf{IOPS} \cite{TSBAD}: a dataset with performance indicators that reflect the scale, quality of web services, and health status of a machine.
    \item \textbf{WSD} \cite{WSD}: a web service dataset, which contains real-world KPIs collected from large Internet companies.
    \item \textbf{LTDB}\cite{LTDB}: a collection of 7 long-duration ECG recordings (14 to 22 hours each), with manually reviewed beat annotations.
    \item \textbf{SVDB} \cite{SVDB}: includes 78 half-hour ECG recordings chosen to supplement the examples of supraventricular arrhythmias in the MIT-BIH Arrhythmia Database.
    \item \textbf{SMAP} \cite{MSL}: real spacecraft telemetry data with anomalies from Soil Moisture Active Passive satellite. It contains time series with one feature representing a sensor measurement, while the rest represent binary encoded commands.
    \item \textbf{MSL} \cite{MSL} : collected from Curiosity Rover on Mars satellite.
\end{itemize}
\subsection{Baselines} \label{sec:apendixbas}
We compare our method against a diverse set of baseline models from both traditional statistical and neural network-based paradigms, selected from the TSB-AD benchmark or manually integrated for fair comparison.  The detailed baselines are described as follows:

\textbf{Traditional Statistical Methods}
\begin{itemize}
    \item \textbf{PCA} \cite{SubPCA} projects data to a lower-dimensional hyperplane, with significant deviation from this plane indicating potential outliers.
    \item \textbf{KNN} \cite{SubKNN} produces the anomaly score of the input instance as the distance to its k-th nearest neighbor.
    \item \textbf{MatrixProfile} \cite{Matrixprofile} identifies anomalies by pinpointing the subsequence exhibiting the most substantial nearest neighbor distance.
    \item \textbf{IForest} \cite{isolationforest} constructs the binary tree, wherein the path length from the root to a node serves as an indicator of anomaly likelihood; shorter paths suggest higher anomaly probability.
    \item \textbf{KShapeAD} \cite{KshapeAD2}  identifies the normal pattern based on the k-Shape clustering algorithm and computes anomaly scores based on the distance between each sub-sequence and the normal pattern. KShapeAD improves KMeansAD as it relies on a more robust time series clustering method and corresponds to an offline version of the streaming SAND method \cite{SAND}.
\end{itemize}

\textbf{Neural Network-based Method}
\begin{itemize}
    \item \textbf{CNN} \cite{CNN} employ Convolutional Neural Network (CNN) to predict the next time stamp on the defined horizon and then compare the difference with the original value.
    \item \textbf{LSTMAD} \cite{LSTMAD} utilizes Long Short-Term Memory (LSTM) networks to model the relationship between current and preceding time series data, detecting anomalies through discrepancies between predicted and actual values.
    \item \textbf{TranAD} \cite{Tranad} is a deep transformer network-based method, which leverages self-conditioning and adversarial training to amplify errors and gain training stability.
    \item \textbf{USAD} \cite{USAD} is based on adversely trained autoencoders, and the anomaly score is the combination of discriminator and reconstruction loss.
    \item \textbf{OmniAnomaly} \cite{OmniAnomaly} is a stochastic recurrent neural network, which captures the normal patterns of time series by learning their robust representations with key techniques such as stochastic variable connection and planar normalizing flow, reconstructs input data by the representations, and use the reconstruction probabilities to determine anomalies.
    \item \textbf{TimesNet} \cite{TimesNet} is a general time series analysis model with applications in forecasting, classification, and anomaly detection. It features TimesBlock, which can discover the multi-periodicity adaptively and extract the complex temporal variations from transformed 2D tensors by a parameter-efficient inception block.
    \item \textbf{DeepSVDD} \cite{deepsvdd} constructs a single-class hypersphere, clustering normal samples inside the sphere and abnormal samples outside the sphere.
    \item \textbf{AnomalyTransformer} \cite{AnomalyTrans} proposes the Anomaly-Attention mechanism, can simultaneously model prior associations and series associations to calculate association discrepancies. 
    \item \textbf{THOC} \cite{THOC} uses dilated recurrent neural network with skip connections to effectively extract multiscale temporal features from time series, and multiple hyperspheres are obtained through a hierarchical clustering process.
    \item \textbf{MEMTO} \cite{MEMTO} proposes a memory-guided transformer, which reduces over-generalization through gated memory modules and incremental training strategies and increases the difficulty of reconstructing abnormal samples by limiting the encoder's ability to handle anomalies, thereby effectively distinguishing normal and abnormal data.
\end{itemize}

\newpage
\subsection{Hyperparameters of Baselines} \label{sec:baspara}
To enhance reproducibility and fairness, we show the hyperparameter settings of the baselines, as shown in the Table \ref{tab:unihyper} and Table \ref{tab:multihyper}.
\begin{table*}[h]
\centering
\begin{tabular}{lcc}
\toprule
\textbf{Method} & \textbf{Hyperparameter 1} & \textbf{Hyperparameter 2} \\
\midrule
PCA          & periodicity: [1, 2, 3]     & n\_components: [0.25, 0.5, 0.75, None] \\
KNN        & periodicity: [1, 2, 3]     & n\_neighbors: [10, 20, 30, 40, 50] \\
MatrixProfile   & periodicity: [1, 2, 3]    & None  \\
IForest      & n\_estimators: [25, 50, 100, 150, 200]     & None     \\
KShapeAD           & periodicity: [1, 2, 3]    & None     \\
CNN          & win\_size: [50, 100, 150]     & num\_channel: [[32,32,40],[16,32,64]] \\
LSTMAD          & win\_size: [50, 100, 150]     & lr: [0.0004, 0.0008]      \\
TranAD          & win\_size: [5, 10, 50]     & lr: [0.001, 0.0001]      \\
USAD          & win\_size: [5, 50, 100]     & lr: [0.001, 0.0001, 1e-05]      \\
TimesNet          & win\_size: [32, 96, 192]     & lr: [0.001, 0.0001, 1e-05]      \\
AnomalyTransformer          & win\_size: [50, 100, 150]     & lr: [0.001, 0.0001, 1e-05]      \\
DeepSVDD          & win\_size: [50, 100, 150]     & lr: [0.001, 0.0001]      \\
THOC          & win\_size: [10, 50, 100]     & dilations: [1, 2, 4] \\
MEMTO          & win\_size: [50, 100]     & lr: [0.001, 0.0001]      \\
\bottomrule
\end{tabular}
\caption{Hyperparameter Settings for Univariate Baselines}
\label{tab:unihyper}
\end{table*}

\begin{table*}[h]
\centering
\begin{tabular}{lcc}
\toprule
\textbf{Method} & \textbf{Hyperparameter 1} & \textbf{Hyperparameter 2} \\
\midrule
PCA          & n\_components: [0.25, 0.5, 0.75, None]     & None \\
KNN          & n\_neighbors: [10,20,30,40,50]     & method: [largest,mean,median] \\
IForest      & n\_estimators: [25, 50, 100, 150, 200]     & max\_features: [0.2, 0.4, 0.6, 0.8, 1.0]     \\
CNN          & win\_size: [50, 100, 150]     & num\_channel: [[32,32,40],[16,32,64]] \\
LSTMAD          & win\_size: [50, 100, 150]     & lr: [0.0004, 0.0008]      \\
AutoEncoder          & win\_size: [50, 100, 150]     & hidden\_neurons: [[64, 32], [32, 16], [128, 64]]      \\
TranAD          & win\_size: [5, 10, 50]     & lr: [0.001, 0.0001]      \\
USAD          & win\_size: [5, 50, 100]     & lr: [0.001, 0.0001, 1e-05]      \\
TimesNet          & win\_size: [32, 96, 192]     & lr: [0.001, 0.0001, 1e-05]      \\
AnomalyTransformer          & win\_size: [50, 100, 150]     & lr: [0.001, 0.0001, 1e-05]      \\
DeepSVDD      & win\_size: [50, 100, 150]     & lr: [0.001, 0.0001]      \\
THOC          & win\_size: [10, 50, 100]     & dilations: [1, 2, 4] \\
MEMTO          & win\_size: [50, 100]     & lr: [0.001, 0.0001]      \\
\bottomrule
\end{tabular}
\caption{Hyperparameter Settings for Multivariate Baselines}
\label{tab:multihyper}
\end{table*}

\newpage
\subsection{Evaluation Metrics}

To comprehensively evaluate the effectiveness of time series anomaly detection models, we adopt three complementary metrics: \textbf{VUS-PR}, \textbf{VUS-ROC}, and \textbf{Affiliation-F1}. These metrics jointly assess ranking quality, temporal tolerance, and segment-level alignment.

\paragraph{VUS-PR.}
VUS-PR is a tolerance-aware generalization of AUC-PR that integrates over both decision thresholds and temporal tolerance levels. It evaluates how well anomaly scores rank true anomalies higher than normal points, while allowing for slight temporal deviations.

For a given temporal tolerance window \(\Delta\), a predicted anomaly at time \(\hat{t}\) is considered correct if there exists a ground-truth anomaly \(t\) such that \(|\hat{t} - t| \leq \Delta\). Under this definition, we compute precision and recall as:
\begin{equation}
\text{TP}_\Delta = \left| \left\{ \hat{t} \in \hat{\mathcal{A}} \,\middle|\, \exists t \in \mathcal{A},\, |\hat{t} - t| \leq \Delta \right\} \right|,
\end{equation}
\begin{equation}
\text{Precision}_\Delta = \frac{\text{TP}_\Delta}{|\hat{\mathcal{A}}|}, \quad
\text{Recall}_\Delta = \frac{\text{TP}_\Delta}{|\mathcal{A}|},
\end{equation}
where \(\hat{\mathcal{A}}\) is the set of predicted anomaly indices, and \(\mathcal{A}\) is the set of ground-truth anomaly indices.

Let \( r \in [0, 1] \) denote the recall value, and \( \text{Precision}_\Delta(r) \) the precision at recall \( r \) under tolerance \(\Delta\). Then, the VUS-PR score is defined as:
\begin{equation}
\text{VUS-PR} = \frac{1}{|\Delta|} \sum_{\Delta_i} \int_0^1 \text{Precision}_{\Delta_i}(r) \, dr,
\end{equation}
this metric is particularly suitable for imbalanced and noisy time series data where perfect temporal alignment is unrealistic.

\paragraph{VUS-ROC.}
VUS-ROC extends the classic AUC-ROC by introducing tolerance to temporal lag, capturing the trade-off between true and false positive rates under relaxed evaluation.

For each tolerance level \(\Delta\), we define:
\begin{itemize}
    \item \( f \in [0, 1] \): false positive rate (FPR),
    \item \( \text{TPR}_\Delta(f) \): true positive rate at FPR \(f\), under tolerance \(\Delta\),
    \item \( |\Delta| \): number of tolerance values considered.
\end{itemize}

The VUS-ROC is computed as the average area under the ROC curve over different tolerance levels:
\begin{equation}
\text{VUS-ROC} = \frac{1}{|\Delta|} \sum_{\Delta_i} \int_0^1 \text{TPR}_{\Delta_i}(f) \, df,
\end{equation}
this metric offers a temporally relaxed ranking view, complementing VUS-PR by emphasizing true negative discrimination.

\paragraph{Affiliation-F1.}
Affiliation-F1 is a range-based metric that evaluates the alignment between predicted anomaly points and ground-truth anomaly intervals. It uses a Gaussian kernel to softly assign affiliation scores based on temporal proximity. Let:
\begin{itemize}
    \item \(\hat{\mathcal{A}}\): predicted anomaly timestamps,
    \item \(\{\mathcal{A}_j\}\): ground truth anomaly intervals,
    \item \(d(x, \mathcal{A}_j)\): temporal distance between point \(x\) and interval \(\mathcal{A}_j\),
    \item \(\sigma\): Gaussian kernel bandwidth.
\end{itemize}

Then, the soft affiliation precision is:
\begin{equation}
\text{Precision}_{\text{Aff}} = \frac{1}{|\hat{\mathcal{A}}|} \sum_{\hat{t} \in \hat{\mathcal{A}}} \max_j \exp\left( -\frac{d(\hat{t}, \mathcal{A}_j)^2}{2\sigma^2} \right),
\end{equation}
similarly, the recall is computed as:
\begin{equation}
\text{Recall}_{\text{Aff}} = \frac{1}{|\mathcal{A}|} \sum_{t \in \mathcal{A}} \max_k \exp\left( -\frac{d(t, \hat{\mathcal{A}}_k)^2}{2\sigma^2} \right),
\end{equation}
the final Affiliation-F1 score is the harmonic mean:
\begin{equation}
\text{Affiliation-F1} = 2 \cdot \frac{\text{Precision}_{\text{Aff}} \cdot \text{Recall}_{\text{Aff}}}{\text{Precision}_{\text{Aff}} + \text{Recall}_{\text{Aff}}},
\end{equation}
this metric is especially effective for evaluating region-level consistency between prediction and annotation, and penalizes predictions that fail to align with any anomaly window.

\section{Model Complexity Analysis}
We analyze the computational complexity of GBOC in terms of its three core components: (1) the encoder, (2) the granular-ball vector data description construction, and (3) the inference process.

The encoder is implemented as a three layer LSTM, which processes each time series subsequence of window length \( w \), input dimension \( d \), and produces a latent representation of dimension \( d' \). The computational complexity of a single LSTM layer is \( \mathcal{O}(w \cdot d' \cdot (d + d')) \), and thus the total complexity across three layers remains linear in \( w \), \( d \), and \( d' \). That is,
\(\mathcal{O}\left(w \cdot d' \cdot (d + d')\right)\). 
This design enables the model to efficiently capture temporal dependencies at low computational cost, particularly when compared to \textit{attention}-based architectures.

\begin{figure}[h!]
    \centering  \includegraphics[width=0.6\linewidth]{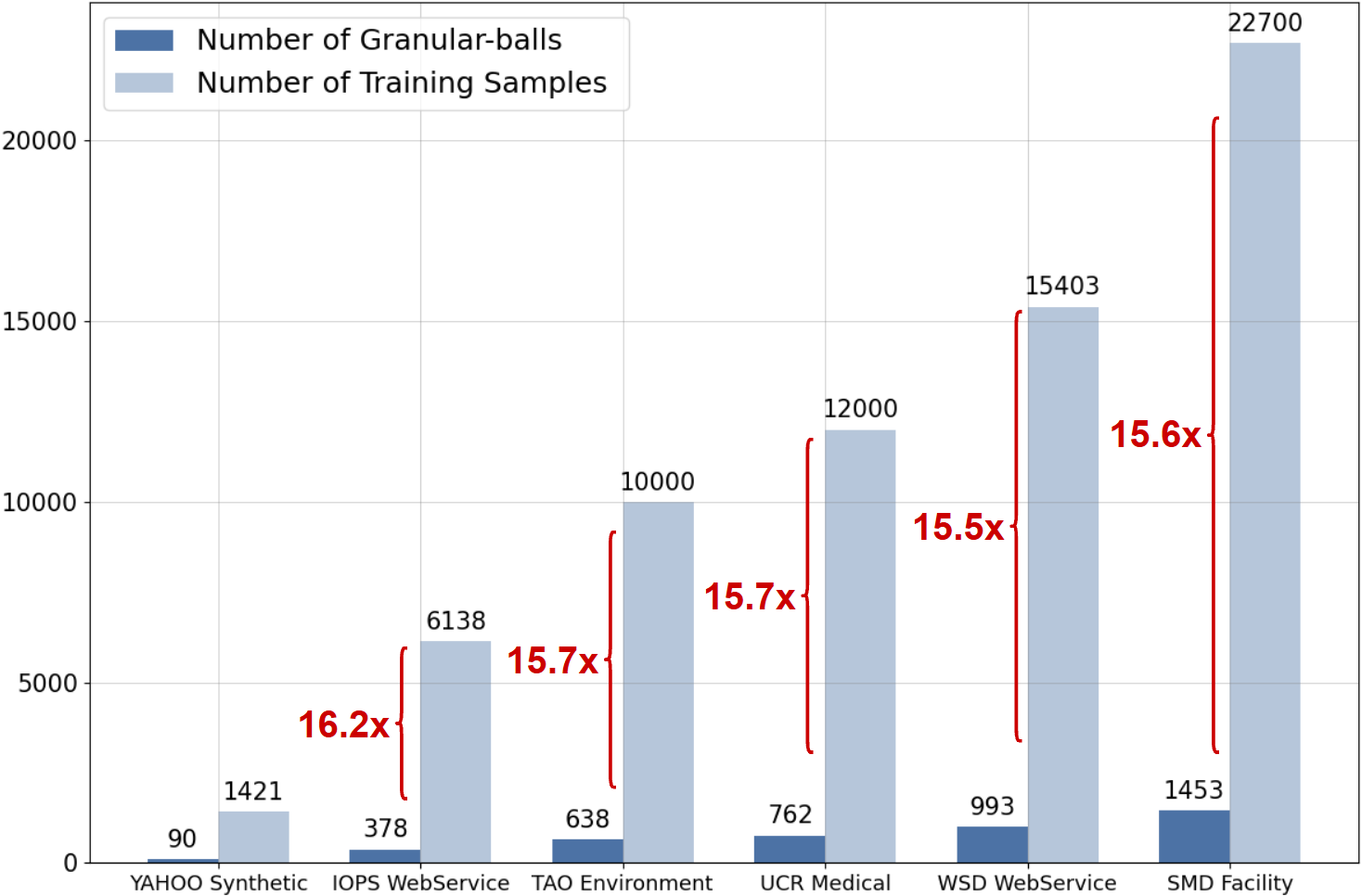}
    \caption{{Granular-balls vs Training Samples}}
    \label{fig:gbvsori}
\end{figure}

To model normal regions in the latent space, GBOC constructs a granular-ball vector data description via a density-guided hierarchical partitioning scheme. This process recursively splits high-density regions into semantically coherent and compact granular-balls, without requiring a predefined number of clusters or global optimization. Since the partitioning is conducted locally and incrementally, its computational cost is approximately
\(\mathcal{O}\left(N_{\text{w}} \log N_{\text{w}}\right)\), 
where \( N_{\text{w}}\) is the number of training subsequences (sliding windows). A post-processing step prunes diffuse or low-confidence granular-balls, yielding a compact memory set. As shown in Figure \ref{fig:gbvsori}, the number of granular-balls generated by this process is consistently much smaller than the number of original training samples across all datasets.

To quantitatively assess the representational quality of the constructed centers, we define a coverage rate metric that can be applied to both granular-ball and K-Means clustering results. Given a set of embedded latent vectors $\mathcal{X} = \{x_i\} $ and a set of centers $\mathcal{C} = \{c_i\}$, the coverage rate is computed as:
\begin{equation}
    \text{Coverage} = \left( 1 - \frac{1}{|\mathcal{X}|} \sum_{x_i \in \mathcal{X}} \frac{\min_{c_j \in C} \| x_i - c_j \|_2}{\max(\mathcal{X}) - \min(\mathcal{X})} \right) \times 100\%,
\end{equation}
This value reflects the average normalized proximity between each latent vector and its closest center, offering an interpretable measure of spatial compactness and memory efficiency.
\begin{table}[t!]
  \centering
  \scalebox{1}{
  \begin{tabular}{c|ccccccc}
    \toprule
    $\mathcal{\text{Method}}$ & SMD    & IOPS   & UCR   & YAHOO  & TAO   & WSD  \\
    \midrule
    $6$-Means & 98.52  & 97.41  & 98.02  & 97.45  & 97.62 & 98.15 \\
    $12$-Means & 99.29  & 98.54  & 98.93  & 98.70  & 98.64 & 98.86 \\
    $Granular$-$ball$ & \textbf{99.92}  & \textbf{99.92}  & \textbf{99.89}  & \textbf{99.70}  & \textbf{99.94} & \textbf{99.88} \\
    \bottomrule
  \end{tabular}
  }
  \caption{Comparison of coverage rates across datasets using Granular-ball and K-Means clustering (with K=6 and K=12)}
  \label{tab: gbvskm} 
\end{table}
To further illustrate the efficiency of the proposed model, we compared GBOC with the popular clustering-based THOC \cite{THOC} in terms of end-to-end training time, total sample inference time, and peak GPU memory usage on the UCR Medical dataset. 

As shown in Table \ref{tab:complexity_thoc}, GBOC has significantly lower training costs and faster inference speed.
\begin{table}[h!]
\centering
\begin{tabular}{lccc}
\toprule
Model & Train Time (s) & Test Time (s) & Peak Memory (GB) \\
\midrule
THOC & 96 & 17.4 & 1.28 \\
\textbf{GBOC}& \textbf{71} & \textbf{11} & \textbf{0.84} \\
\bottomrule
\end{tabular}
\caption{Runtime and memory comparison between GBOC and THOC under the same setting.}
\label{tab:complexity_thoc}
\end{table}

At test time, each incoming subsequence is encoded into a latent vector and compared against the centers of the retained granular-balls. Since the number of granular-balls \( M \) is typically much smaller than the number of training windows (i.e., \( M \ll N_{\text{w}} \)), the inference cost is dominated by a nearest-neighbor search over \( M \) vectors in the latent space: \(\mathcal{O}(M \cdot d')\). The low number of representative units ensures fast inference, making GBOC suitable for real-time or streaming anomaly detection. At the same time, as detailed in Table \ref{tab: gbvskm}, our experiments also show that GBOC consistently achieves higher coverage than K-Means with fixed cluster counts (e.g., K = 6, 12), validating the effectiveness of its adaptive data description strategy.

In summary, GBOC achieves a favorable trade-off between representational flexibility and computational efficiency. Training complexity is dominated by encoder computation and local partitioning, while inference remains lightweight due to the compact and expressive granular-ball vector data description. The comparison with THOC further confirms that the proposed granular-ball data description provides not only better geometric adaptability but also lower runtime overhead in practice.

\section{Detailed Experimental Results}

\subsection{Comparison Results with Existing GB-based Methods}
To complement our main results, we additionally compare GBOC with two recent granular-ball–based anomaly detection methods, GBMOD \cite{GBMOD} and GBDO \cite{GBDO}. Since these methods were originally developed for static tabular data rather than sequential settings, we apply them directly to window-level latent representations for a fair comparison. We report their performance on six benchmark time-series datasets to illustrate how classical GBC variants behave under temporal anomaly detection scenarios.
\begin{table*}[h!]
  \centering
  \scalebox{0.74}{
    \begin{tabular}{c|ccc|ccc|ccc|ccc|ccc|ccc}
    \toprule
    \multirow{2}[4]{*}{Models} & \multicolumn{3}{c|}{SMD Facility} & \multicolumn{3}{c|}{TAO Environment} & \multicolumn{3}{c|}{YAHOO Synthetic} & \multicolumn{3}{c|}{UCR Medical} & \multicolumn{3}{c|}{IOPS WebService} & \multicolumn{3}{c}{WSD WebService} \\
\cmidrule{2-19}          & VP    & VR    & AF    & VP    & VR    & AF    & VP    & VR    & AF    & VP    & VR    & AF    & VP    & VR    & AF   & VP    & VR    & AF\\
    \midrule
    GBMOD   & 0.510  & 0.845  & 0.931  & 0.905  & 0.968  & 0.129   & 0.221  & 0.625  & 0.586   & 0.810  & 0.921  & 0.921   & 0.221  & 0.625  & 0.583  & 0.301  & 0.801  & 0.782 \\
    GBDO   & 0.598  & 0.878  & 0.941  & 0.928  & 0.975  & 0.134  & 0.285  & 0.701  & 0.608   & 0.882  & 0.955  & 0.930  & 0.265  & 0.656  & 0.592  & 0.355  & 0.833  & 0.798 \\
    \midrule
    \textbf{GBOC}  & \textbf{0.831}  & \textbf{0.999}  & \textbf{0.999}  & \textbf{0.978}  & \textbf{0.991}  & \textbf{0.219}  & \textbf{0.991} & \textbf{1.000} & \textbf{0.950} & \textbf{0.996} & \textbf{0.999} & \textbf{0.996}  & \textbf{0.604}  & \textbf{0.992}  & \textbf{0.948}  & \textbf{0.963}  & \textbf{0.998}  & \textbf{0.995} \\
    \bottomrule
    \end{tabular}
    }
    \caption{Comparison of GBOC with granular-ball–based baselines GBMOD and GBDO on six benchmark time-series datasets. (VP: VUS-PR, VR: VUS-ROC, AF: Affiliation-F1).}
    \label{tab:gbbase}
\end{table*}

As shown in Table \ref{tab:gbbase}, GBOC achieves the best VUS-PR, VUS-ROC, and Affiliation-F1 across all six datasets, showing clear advantages over granular-ball baselines. GBMOD and GBDO perform reasonably on simpler datasets such as TAO Environment, but degrade notably on more challenging cases involving noise or distribution shift, including YAHOO Synthetic, IOPS WebService, and WSD WebService. In contrast, GBOC maintains strong performance in all settings, demonstrating its superior ability to model complex time-series structures.

\clearpage
\subsection{Ablation Studies} \label{sec:apendixabl}
We conduct comprehensive ablation studies to evaluate the effect of GBOC’s core components, including granular-ball computing, the pruning strategy, and its two loss terms: the reconstruction loss $\mathcal{L}_\text{rec}$ and the granular-ball alignment loss $\mathcal{L}_\text{gb}$, Table \ref{tab: con-ROC}, Table \ref{tab: con-AF}, Table \ref{tab: Loss-VR}, Table \ref{tab: Loss-AF} present the results using VUS-ROC and Affiliation-F1 across six representative datasets.
\begin{table}[h!]
  \centering
  \scalebox{1}{
  \begin{tabular}{cc|ccccccc}
    \toprule
    $\mathcal{\text{GBC}}$ & $\mathcal{\text{Pruning}}$ & SMD    & IOPS   & UCR   & YAHOO  & TAO   & WSD  \\
    \midrule
    $\times$ & $\times$ & 0.959  & 0.973  & 0.920  & 0.976  & 0.969 & 0.968 \\
    $\checkmark$     & $\times$ & 0.971  & 0.957  & 0.953  & 0.965  & 0.980 & 0.857 \\
    $\checkmark$ & $\checkmark$ & \textbf{0.999}  & \textbf{0.992}  & \textbf{0.999}  & \textbf{1.000}  & \textbf{0.991} & \textbf{0.998} \\
    \bottomrule
  \end{tabular}
  }
  \caption{Effects of \textit{w/o granular-ball computing} and \textit{w/o pruning} regarding VUS-ROC.}
  \label{tab: con-ROC} 
\end{table}
\begin{table}[h!]
  \centering
  \scalebox{1}{
  \begin{tabular}{cc|ccccccc}
    \toprule
    $\mathcal{\text{GBC}}$ & $\mathcal{\text{Pruning}}$ & SMD    & IOPS   & UCR   & YAHOO  & TAO   & WSD  \\
    \midrule
    $\times$ & $\times$ & 0.975  & 0.891  & 0.886  & 0.891  & 0.120 & 0.952 \\
    $\checkmark$     & $\times$ & 0.921  & 0.837  & 0.920  & 0.743  & 0.149 & 0.964 \\
    $\checkmark$ & $\checkmark$ & \textbf{0.999}  & \textbf{0.948}  & \textbf{0.996}  & \textbf{0.950}  & \textbf{0.219} & \textbf{0.995} \\
    \bottomrule
  \end{tabular}
  }
  \caption{Effects of \textit{w/o granular-ball computing} and \textit{w/o pruning} regarding Affiliation-F1.}
  \label{tab: con-AF} 
\end{table}
\begin{table}[h!]
  \centering
  \scalebox{1}{
  \begin{tabular}{cc|ccccccc}
    \toprule
    $\mathcal{L}_{\text{rec}}$ & $\mathcal{L}_{\text{gb}}$ & SMD    & IOPS   & UCR   & YAHOO  & TAO   & WSD  \\
    \midrule
    $\checkmark$ & $\times$ & 0.991  & 0.986  & 0.953  & 0.928  & 0.989 & 0.978 \\
    $\times$     & $\checkmark$ & 0.993  & 0.985  & 0.946  & 0.947  & 0.989 & 0.980 \\
    $\checkmark$ & $\checkmark$ & \textbf{0.999}  & \textbf{0.992}  & \textbf{0.999}  & \textbf{1.000}  & \textbf{0.991} & \textbf{0.998} \\
    \bottomrule
  \end{tabular}
  }
  \caption{Effectiveness of $\mathcal{L}_{\text{rec}}$ and $\mathcal{L}_{\text{gb}}$ on VUS-ROC.}
  \label{tab: Loss-VR} 
\end{table}
\begin{table}[h!]
  \centering
  \scalebox{1}{
  \begin{tabular}{cc|ccccccc}
    \toprule
    $\mathcal{L}_{\text{rec}}$ & $\mathcal{L}_{\text{gb}}$ & SMD    & IOPS   & UCR   & YAHOO  & TAO   & WSD  \\
    \midrule
    $\checkmark$ & $\times$ & 0.887  & 0.889  & 0.905  & 0.851  & 0.126 & 0.970 \\
    $\times$     & $\checkmark$ & 0.765  & 0.869  & 0.932  & 0.819  & 0.127 & 0.814 \\
    $\checkmark$ & $\checkmark$ & \textbf{0.999}  & \textbf{0.948}  & \textbf{0.996}  & \textbf{0.950}  & \textbf{0.219} & \textbf{0.995} \\
    \bottomrule
  \end{tabular}
  }
  \caption{Effectiveness of $\mathcal{L}_{\text{rec}}$ and $\mathcal{L}_{\text{gb}}$ on Affiliation-F1.}
  \label{tab: Loss-AF} 
\end{table}

Table \ref{tab: con-ROC} and Table \ref{tab: con-AF} show that removing either granular-ball computing or pruning consistently leads to performance degradation. The absence of granular-ball computing results in the most significant drop, especially on challenging datasets such as UCR and YAHOO. This highlights its importance in modeling local data structure and maintaining alignment with the latent distribution. The pruning strategy provides additional gains by eliminating low-quality or ambiguous regions, which is particularly beneficial for datasets with noise or label imbalance, such as YAHOO and WSD.

Table \ref{tab: Loss-VR} and Table \ref{tab: Loss-AF} examine the contributions of the two loss components. Removing either $\mathcal{L}_\text{rec}$ and $\mathcal{L}_\text{gb}$ results in a noticeable drop in both evaluation metrics, especially in Affiliation-F1. This indicates that both losses play essential roles. The reconstruction loss preserves temporal fidelity by encouraging faithful reconstruction of the input, while the alignment loss enforces compactness in the latent space by guiding samples toward the centers of their corresponding granular-balls. Their combination leads to the best performance, suggesting that both structural preservation and discriminative alignment are crucial for effective anomaly detection.

\clearpage
\subsection{Hyperparameter Sensitivity Analysis} \label{sec:apendixhyper}
\begin{figure*}[h]
    \centering  \includegraphics[width=0.9\linewidth]{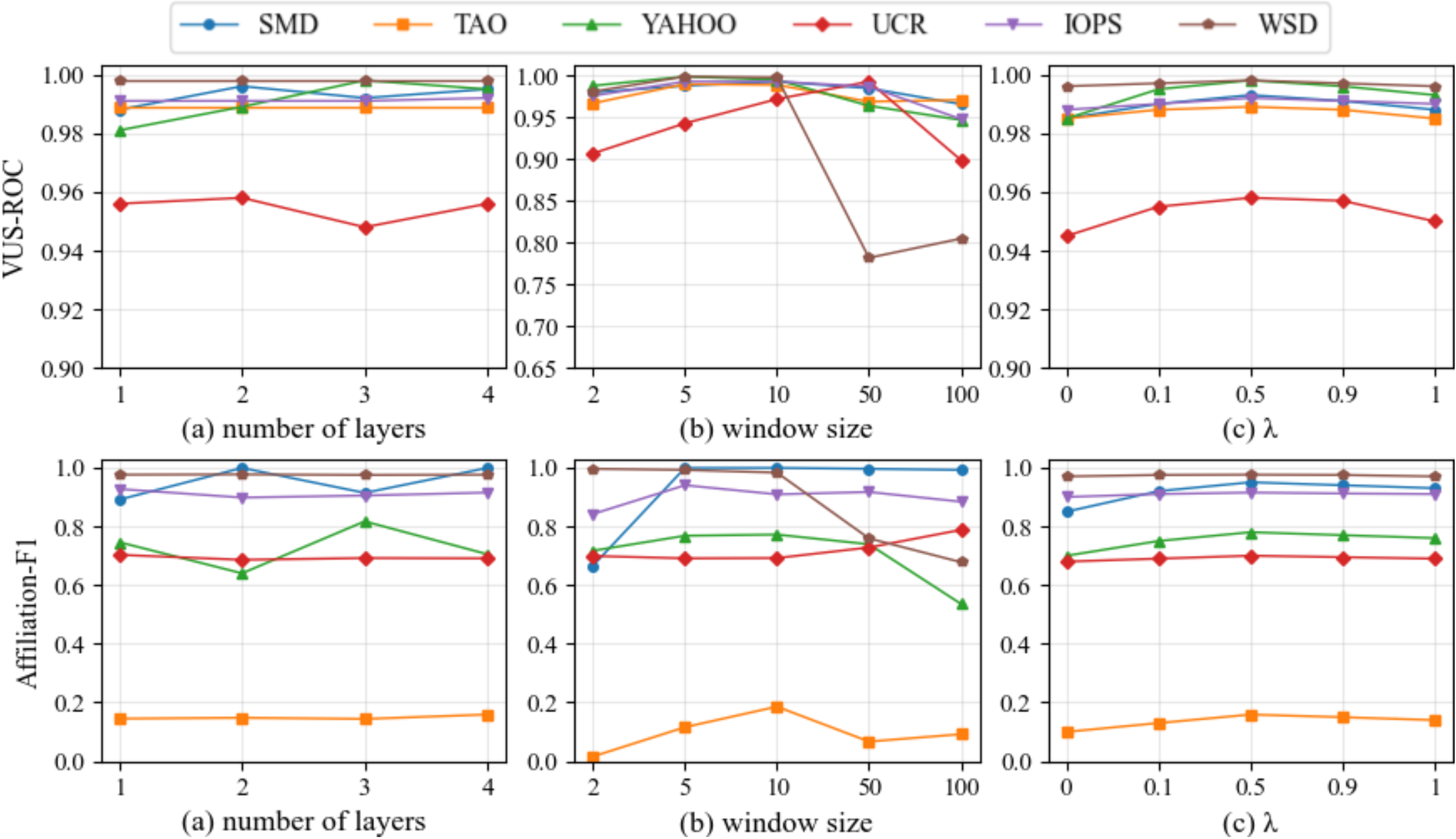}
    \caption{{Sensitivity analysis of GBOC on VUS-ROC and Affiliation-F1 to the number of LSTM encoder layers, input window size and loss weight $\lambda$ across multiple datasets.}}
    \label{fig:hypervusaf}
\end{figure*}

We conduct a comprehensive sensitivity analysis on its three key hyperparameters: the window size, the number of LSTM encoder layers, and the loss weight $\lambda$. We visualize their impact on detection performance using line plots of VUS-ROC and Affiliation-F1 across multiple datasets. As shown in Figure~\ref{fig:hypervusaf} (left), the number of LSTM layers is a relatively insensitive hyperparameter, with a two or three-layer configuration generally achieving strong and stable performance across datasets. In Figure~\ref{fig:hypervusaf} (middle), the sensitivity to window size varies across datasets: smaller windows are more effective on the YAHOO dataset, while a larger window size, such as 50 or 100, performs better on UCR. Finally, Figure~\ref{fig:hypervusaf} (right) shows that tuning the loss weight $\lambda$ yields better results than extreme values ($\lambda = 0$ or $\lambda = 1$), validating the importance of balancing structural alignment and reconstruction in our objective function.

\clearpage
\section{More Visualization Results} \label{apendixvis}
\subsection{Visual Diagnosis of Metric Discrepancies} \label{sec:apendixf1nan}
\begin{figure*}[h]
    \centering  \includegraphics[width=0.95\linewidth]{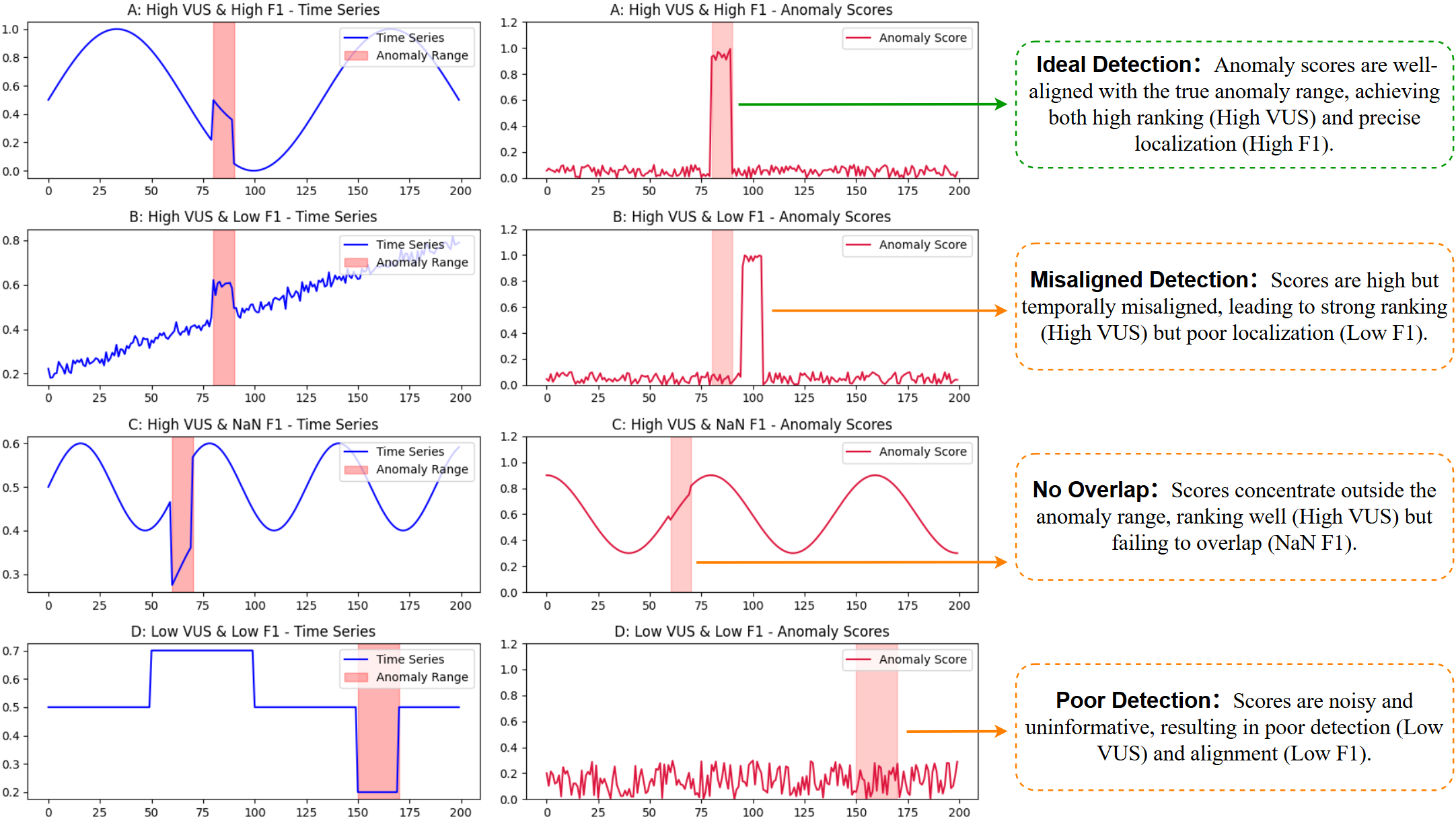}
    \caption{{Example of anomaly score visualization under different evaluation behaviors. Despite achieving high VUS scores in scenarios (B) and (C), the predicted anomaly scores are not temporally aligned with the ground truth intervals, resulting in low or undefined Affiliation-F1. }}
    \label{fig:AffNaN2}
\end{figure*}

To further understand the inconsistency between high VUS-based metrics and low or NaN Affiliation-F1 scores, we perform a detailed visual analysis of representative scenarios. As shown in Figure \ref{fig:AffNaN2}, by comparing the anomaly score curves and their alignment with ground truth anomaly ranges, we identify common failure patterns that lead to metric divergence such as prediction shifts, score inflation outside anomaly windows, and insufficient overlap with annotated ranges. These visualizations reveal that high VUS can occur even when the model fails to localize anomalies precisely, highlighting the importance of range-aware metrics like Affiliation-F1 for evaluating temporal alignment and interpretability in time series anomaly detection.

\clearpage
\subsection{Visualization of Anomaly Detection} \label{sec:apendixad}
\begin{figure}[htbp]
    \centering
    \begin{subfigure}{\linewidth}
        \centering
        \includegraphics[width=0.92\linewidth]{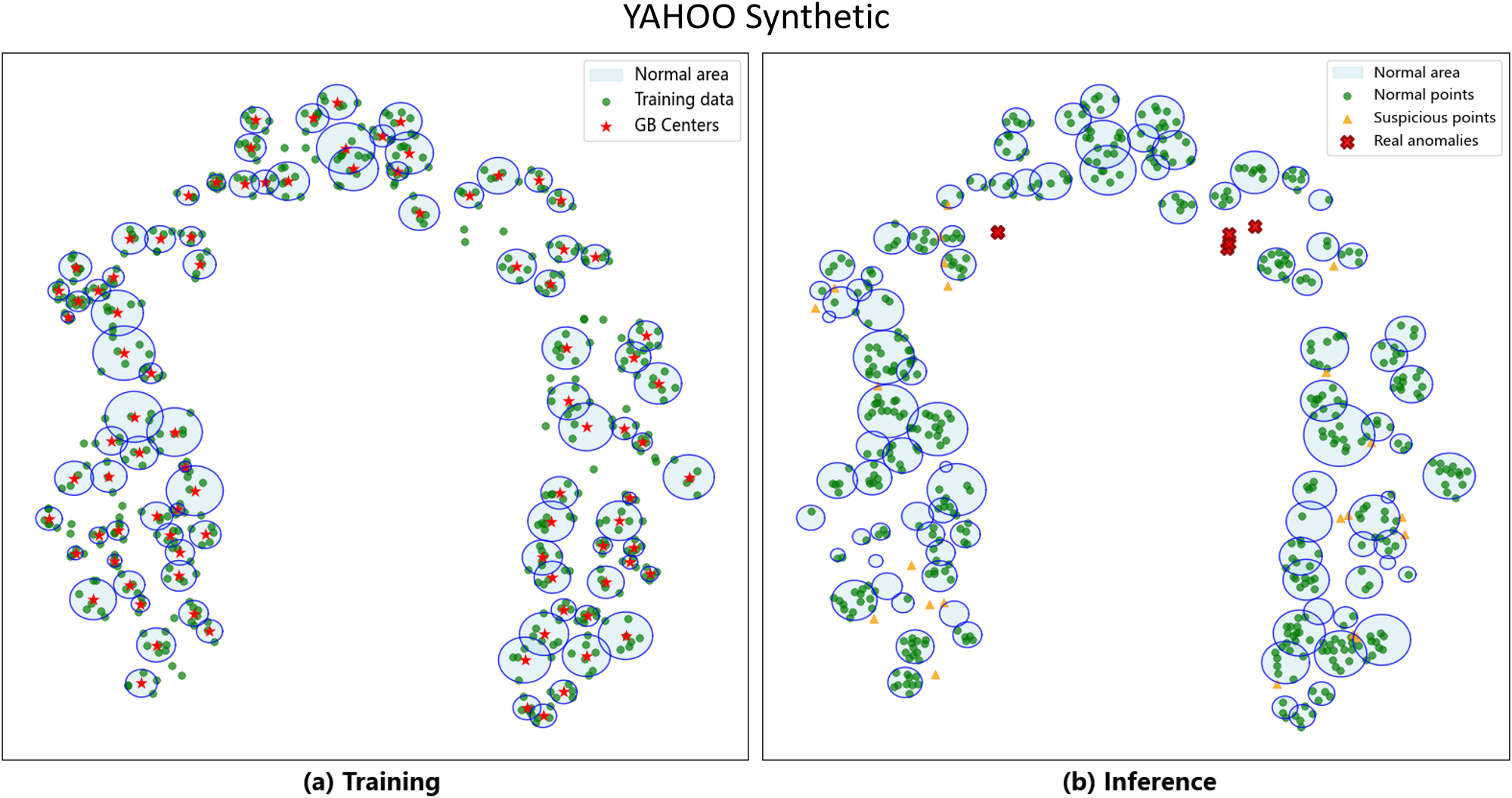}
    \end{subfigure}
    \begin{subfigure}{\linewidth}
        \centering
        \includegraphics[width=0.92\linewidth]{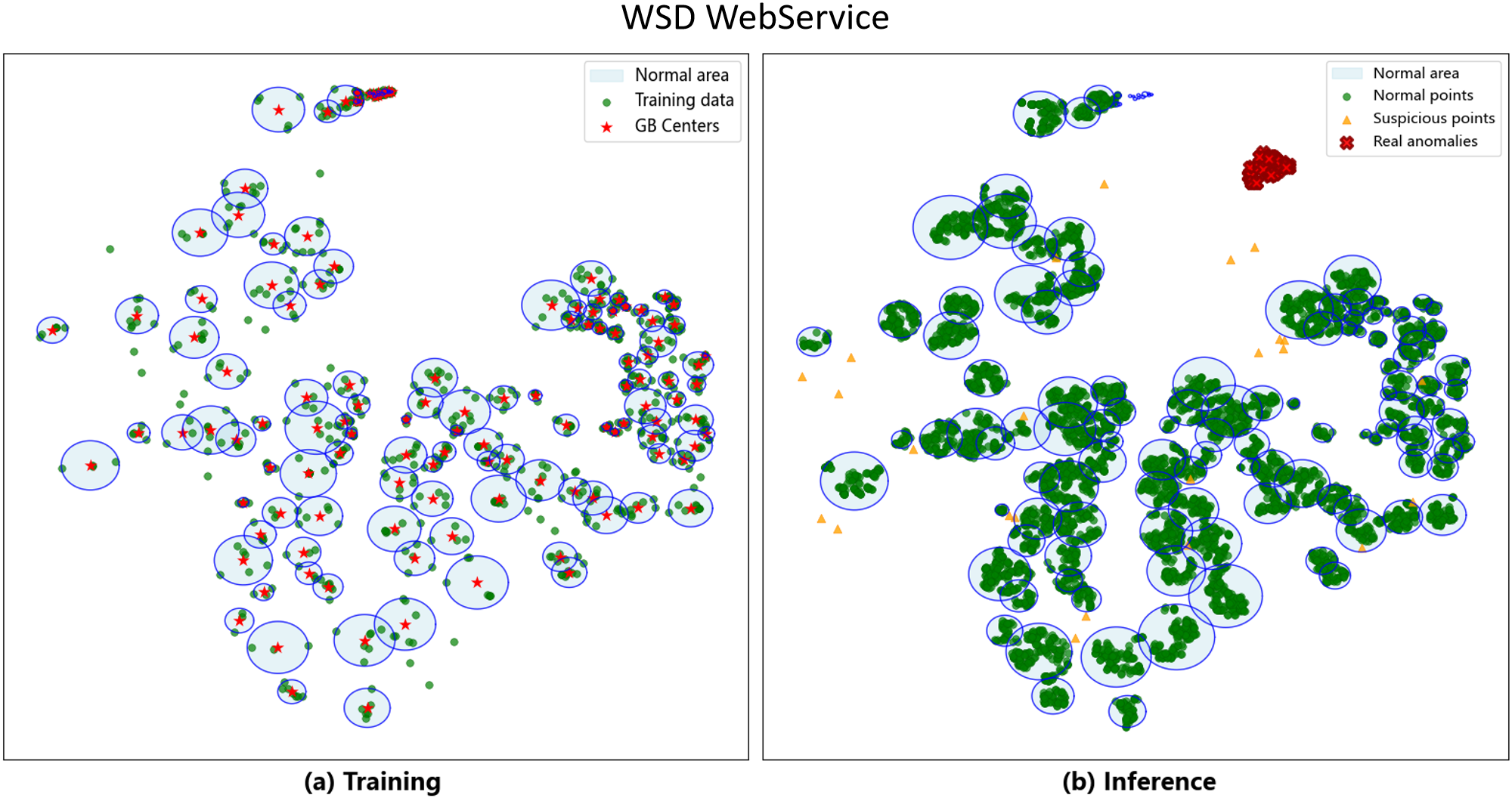}
    \end{subfigure}
    \caption{t-SNE visualization of GBOC's latent representations on multiple datasets.}
    \label{fig:GBOCAD}
\end{figure}

To qualitatively evaluate the generalization and adaptability of GBOC across diverse time series datasets, we visualize its learned latent representations using t-SNE. As shown in Figure \ref{fig:GBOCAD}. Each plot highlights how GBOC organizes normal and anomalous samples in the latent space: normal samples form dense, well-separated granular-ball regions, while anomalies tend to lie outside or on the fringes of these clusters. Across datasets with different characteristics, GBOC consistently preserves structural compactness and effectively isolates abnormal patterns, demonstrating its robustness and interpretability across domains.
\begin{figure}[htbp]
    \centering
    \begin{subfigure}{\linewidth}
        \centering
        \includegraphics[width=0.92\linewidth]{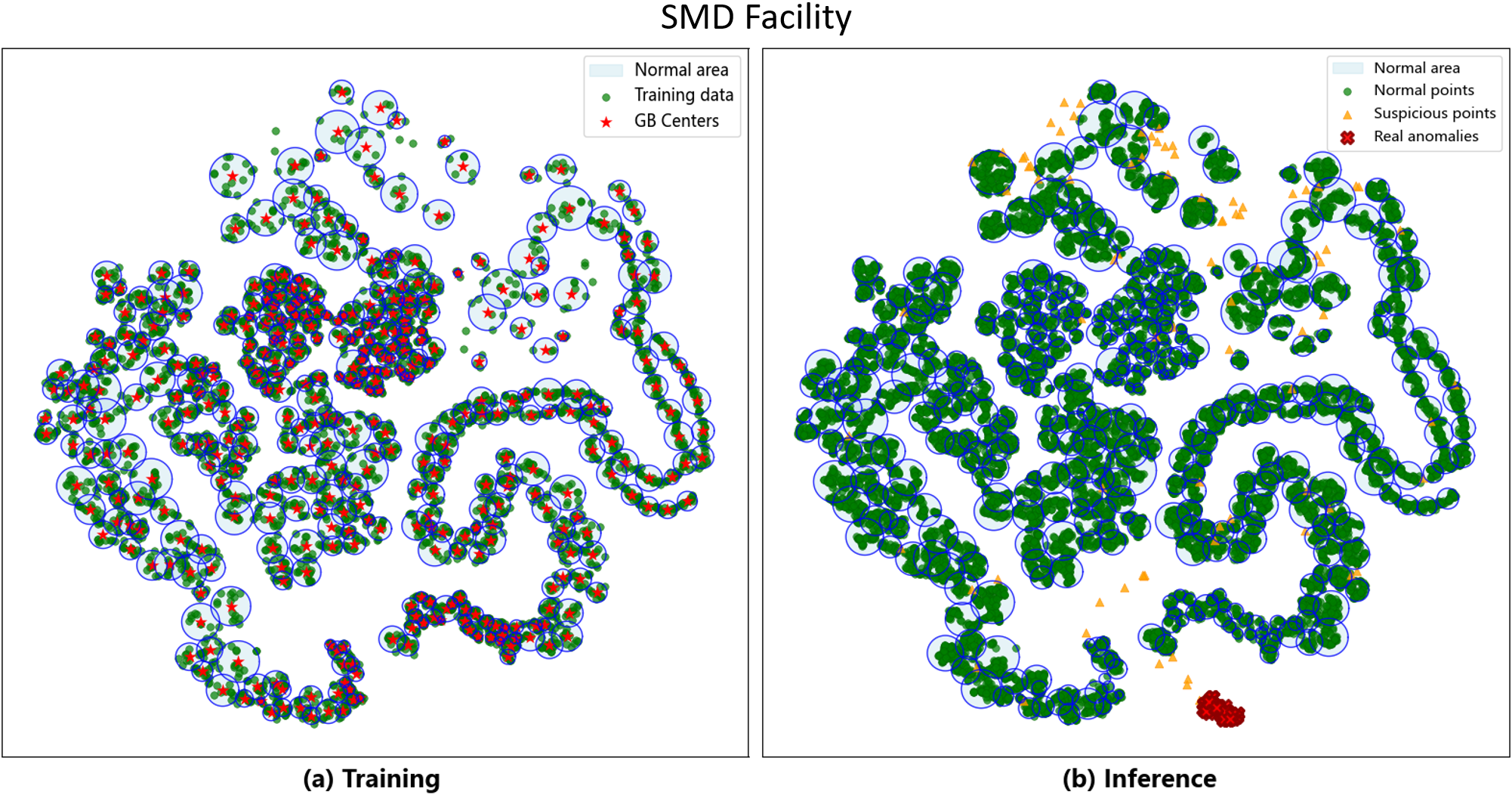}
    \end{subfigure}
    \begin{subfigure}{\linewidth}
        \centering
        \includegraphics[width=0.92\linewidth]{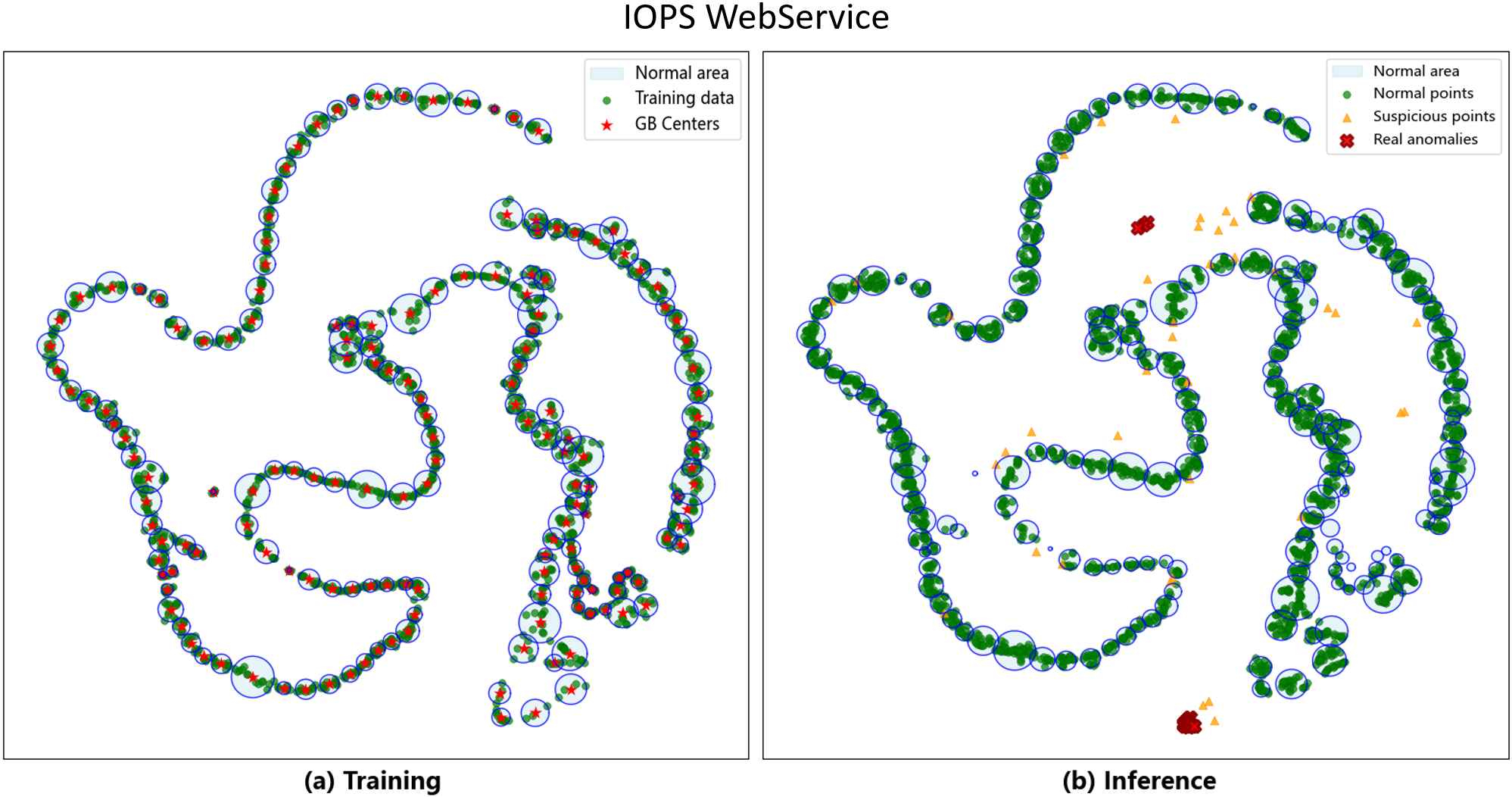}
    \end{subfigure}
\end{figure}

\clearpage
\subsection{Visualization of Anomaly Score} \label{sec:apendixas}

\subsubsection{Comparative Visualization with Related Baselines}
\begin{figure*}[h]
    \centering  \includegraphics[width=0.95\linewidth]{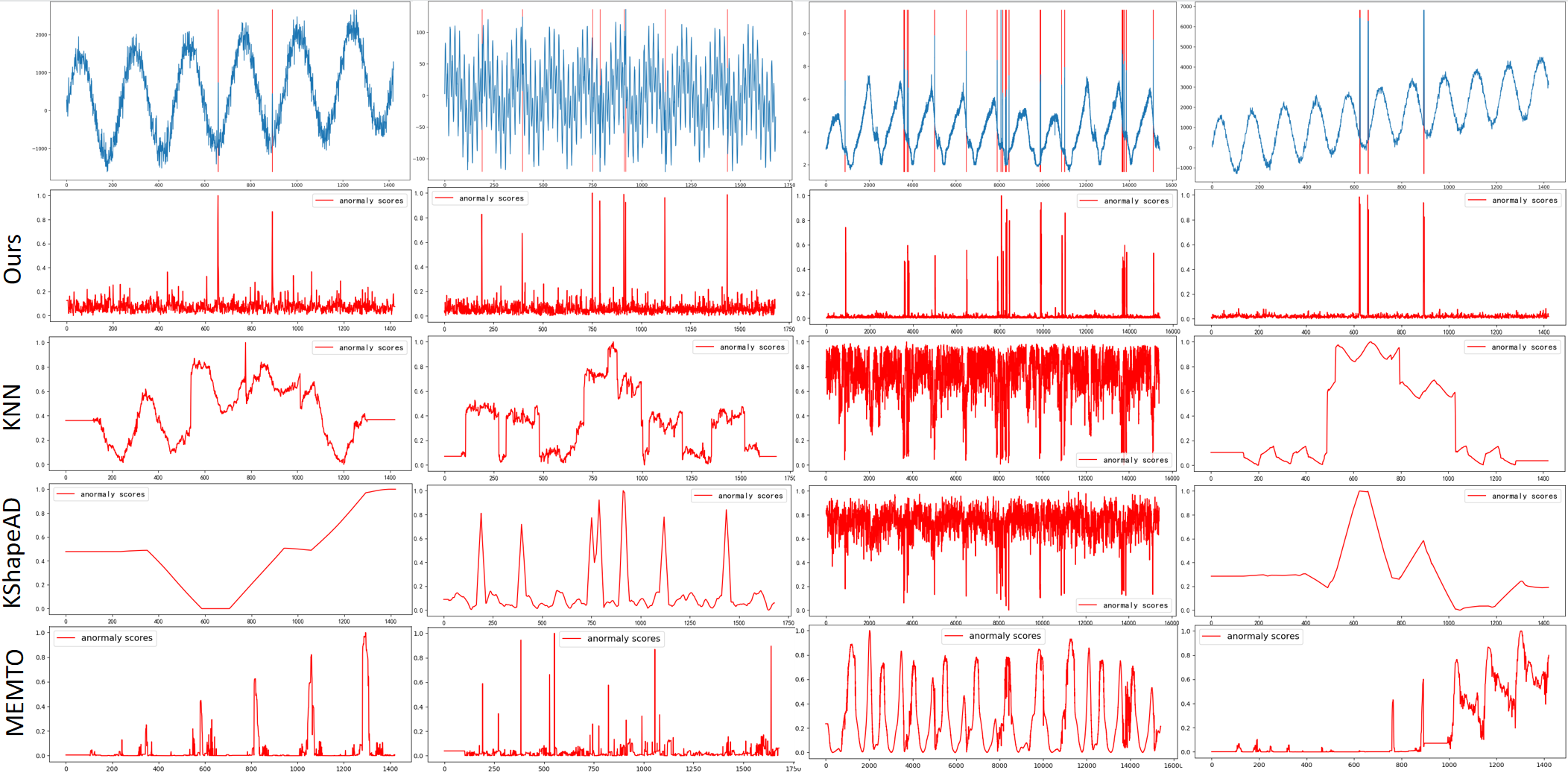}
    \caption{{Comparative visualization of anomaly scores produced by GBOC and baseline methods (KNN, KShapeAD, and MEMTO) from YAHOO and WSD datasets.}}
    \label{fig:viscom}
\end{figure*}

As showm in Figure \ref{fig:viscom}, we present a side-by-side comparison of anomaly scores produced by GBOC and three representative baseline methods, KShapeAD, MEMTO, and KNN. These visualizations reveal that while baseline methods often produce noisy or misaligned anomaly scores in the presence of drift or noise, GBOC consistently generates sharp, well-localized anomaly peaks, demonstrating superior robustness and temporal precision.

\clearpage
\subsubsection{Anomaly Score Distribution by GBOC}
\begin{figure*}[h]
    \centering  \includegraphics[width=0.95\linewidth]{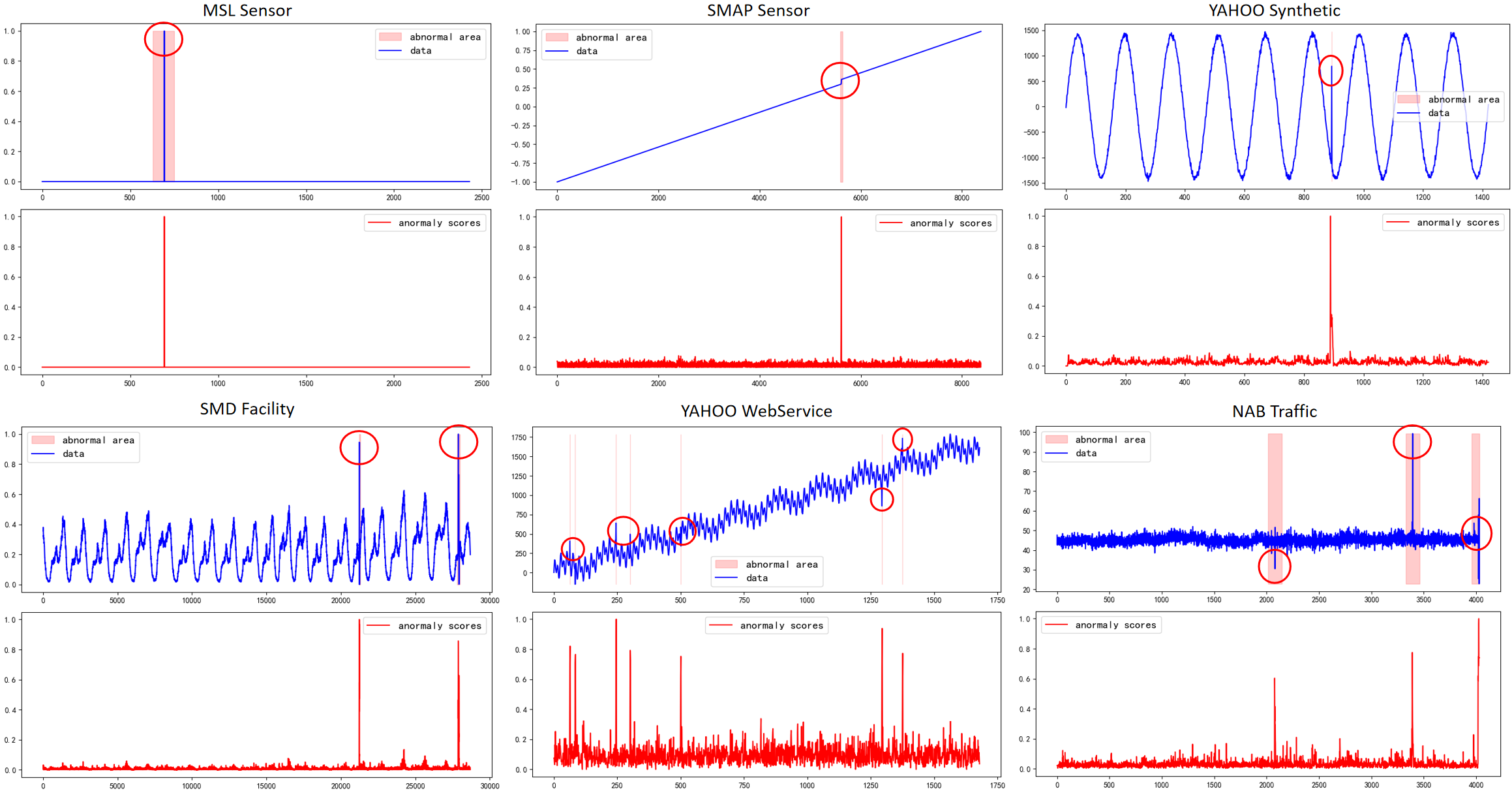}
    \caption{{Anomaly score curves generated by GBOC on multiple datasets.}}
    \label{fig:GBADscore}
\end{figure*}

As showm in Figure \ref{fig:GBADscore}, we further visualize the anomaly scores produced solely by GBOC across multiple datasets. These visualizations highlight the model’s ability to accurately identify true anomaly intervals. The smooth and well-structured score curves reflect the effectiveness of GBOC’s granular-ball representation in modeling normal patterns and isolating deviations.

\end{document}